\documentclass[10pt,twocolumn,letterpaper]{article}

\usepackage[pagenumbers]{cvpr} % To force page numbers, e.g. for an arXiv version
\usepackage{amsmath}
\usepackage{multirow}
\usepackage{pifont}
\usepackage{makecell}
\usepackage{subcaption}
\usepackage{graphicx}
\usepackage{colortbl} % For command \arrayrulecolor
\usepackage{wrapfig}
\usepackage[accsupp]{axessibility}
\usepackage[subtle]{savetrees}

\newcommand{\methodname}[1]{\textcolor{blue}{\texttt{SAGA}}}

\newcommand{\topcolor}{blue!10}
\newcommand{\bottomcolor}{red!10}

\definecolor{creamclr}{HTML}{FFF8E1}   % warm cream
\definecolor{lightblueclr}{HTML}{E3F2FD}    % light blue
\definecolor{lightgreenclr}{HTML}{C8E6C9}      % light green for "ours best"
\newcommand{\creamclr}{creamclr}
\newcommand{\lightblueclr}{lightblueclr}

\newcommand{\fulldata}[1]{\textcolor{red}{\textbf{#1}}}
\newcommand{\oursetting}[1]{\textcolor{blue}{\textbf{#1}}}

\newcommand{\binarylevel}{\textcolor{violet}{\texttt{BIN-L}}}
\newcommand{\tasklevel}{\textcolor{olive}{\texttt{TASK-L}}}
\newcommand{\sdlevel}{\textcolor{cyan}{\texttt{SD-L}}}
\newcommand{\teamlevel}{\textcolor{purple}{\texttt{TEAM-L}}}
\newcommand{\generatorlevel}{{\color[HTML]{FF8C00}{\texttt{GEN-L}}}}

\newcommand{\temporalsign}[1]{\textcolor{blue}{\texttt{T-Sig#1}}}

\definecolor{cvprblue}{rgb}{0.21,0.49,0.74}
\usepackage[pagebackref,breaklinks,colorlinks,allcolors=cvprblue]{hyperref}

\def\paperID{****} % *** Enter the Paper ID here
\def\confName{CVPR}
\def\confYear{2026}

\title{\methodname{}: Source Attribution of Generative AI Videos\footnotemark[1]}

\author{Rohit Kundu$^{1,2}$, Vishal Mohanty$^{2}$, Hao Xiong$^{3}$, Shan Jia$^{2}$, Athula Balachandran$^{2}$, Amit K. Roy-Chowdhury$^{1}$ \\
{$^1$University of California, Riverside, $^2$YouTube (Google), $^3$Google DeepMind}\\
{\tt\small \{rohit.kundu@email, amitrc@ece\}.ucr.edu; \{rohitkun, vishalmohanty, haoxg, shanjia, athula\}@google.com}
}

\begin{document}

%%%%begin teaser fig%%%%
\twocolumn[{%
\renewcommand\twocolumn[1][]{#1}%
\maketitle
\begin{center}
    \centering
    \includegraphics[width=\textwidth]{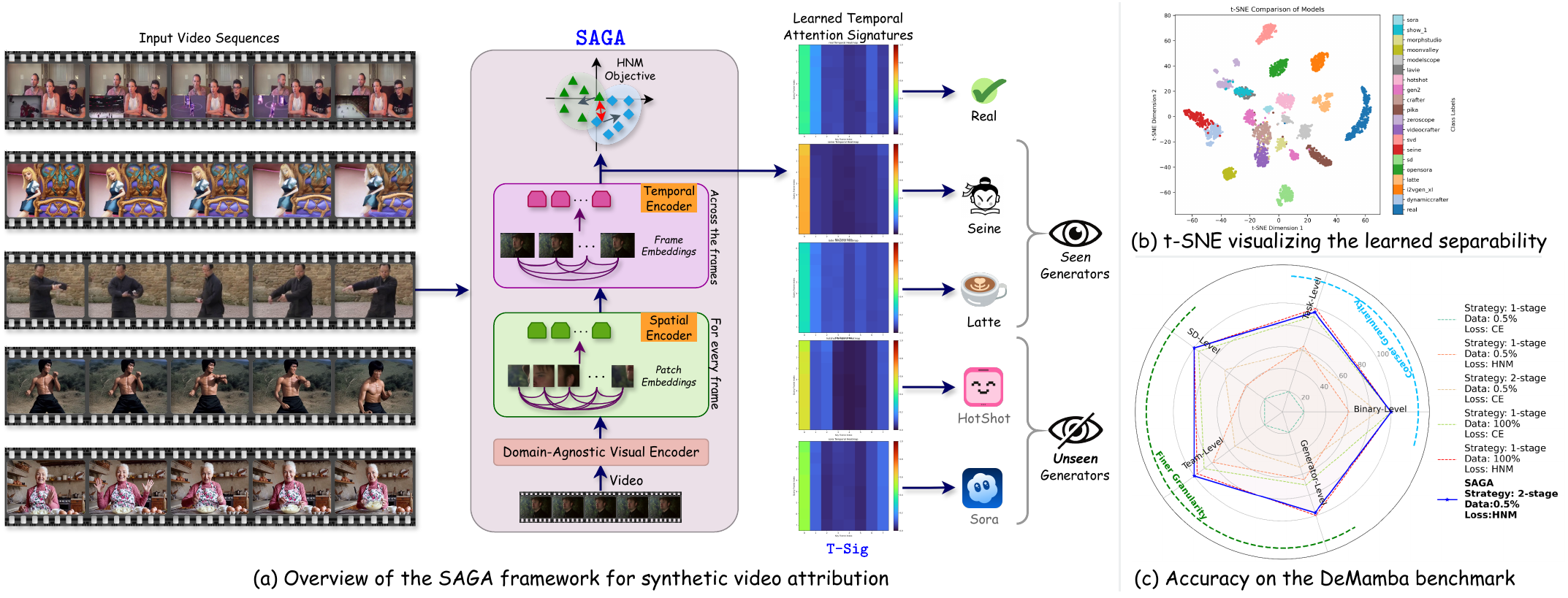}
    \vspace{-2em}
    \captionof{figure}{\textbf{\methodname{}: Data-Efficient \& Interpretable AI Video Source Attribution.} (a) Temporal Attention Signatures (\temporalsign{s}): \methodname{} pioneers AI video source attribution. Our novel \temporalsign{s} provide interpretability, showing unique fingerprints for Real, Seen, and even Unseen generators. (b) Feature Separability: t-SNE visualization of learned features demonstrates clear generator clusters. (c) Multi-Granular Performance \& Data Efficiency: \methodname{} excels across 5 attribution levels. Radar chart shows our 2-stage training method using the Hard Negative Mining (HNM) objective, using only 0.5\% labeled data, matches fully supervised performance and surpasses baselines.}
    \label{fig:teaser}
\end{center}%
}]
\footnotetext[1]{This paper has been published at The IEEE/CVF Conference on Computer Vision and Pattern Recognition (CVPR) 2026}

\begin{abstract}
  The proliferation of generative AI has led to hyper-realistic synthetic videos, escalating misuse risks and outstripping binary real/fake detectors. We introduce \methodname{} (\underline{S}ource \underline{A}ttribution of \underline{G}enerative \underline{A}I videos), the first comprehensive framework to address the urgent need for AI-generated \textit{video source attribution} at a large scale. Unlike traditional detection, \methodname{} identifies the specific generative model used. It uniquely provides multi-granular attribution across five levels: authenticity, generation task (e.g., T2V/I2V), model version, development team, and the precise generator, offering far richer forensic insights. Our novel video transformer architecture, leveraging features from a robust vision foundation model, effectively captures spatio-temporal artifacts. Critically, we introduce a data-efficient pretrain-and-attribute strategy, enabling \methodname{} to achieve state-of-the-art attribution using only 0.5\% of source-labeled data per class, matching fully supervised performance. Furthermore, we propose Temporal Attention Signatures (\temporalsign{s}), a novel interpretability method that visualizes learned temporal differences, offering the first explanation for \textit{why} different video generators are distinguishable. Extensive experiments on public datasets, including cross-domain scenarios, demonstrate that \methodname{} sets a new benchmark for synthetic video provenance, providing crucial, interpretable insights for forensic and regulatory applications. The project page is \href{https://rohit-kundu.github.io/SAGA}{https://rohit-kundu.github.io/SAGA}.
\end{abstract}

\section{Introduction}\label{sec:intro}
The rapid advancement of AI-driven video synthesis, spanning text-to-video (T2V) \cite{wang2023modelscope,opensora,morphstudio} and image-to-video (I2V) \cite{blattmann2023stable,xing2024dynamicrafter,chen2023seine} systems, has democratized content creation but also heightened concerns over misuse and misinformation \cite{sora2024, lin2025omnihuman}, exemplified by incidents like AI-generated wildfire videos causing public alarm \cite{reuters2025wildfires}. Current defenses largely focus on binary real/fake detection \cite{chen2024demamba,kundu2025towards,xu2023tall,cheng2024can}. However, as generative models multiply and evolve at an unprecedented pace \cite{li2024fakebench, lee2024tug}, merely detecting a video as synthetic is insufficient. The critical need has \textbf{shifted from \textit{whether} it's fake to \textit{what is its source}}? Identifying the specific generative model or family, called source attribution \cite{wang2023did, yang2022deepfake, vahdati2024beyond}, is paramount for effective digital forensics \cite{amerini2024deepfake}, intellectual property enforcement \cite{ballardini2019ai,smits2022generative}, and developing robust adversarial countermeasures \cite{pinhasov2024xai, zhou2022adversarial}.

Attributing synthetic videos to their source is a far more complex challenge than traditional DeepFake detection or even image source attribution \cite{wang2023did,yang2022deepfake, fang2023open}. While image-based methods offer a starting point, they fundamentally fail to address video-specific complexities. We identify three key barriers: (1) \textbf{Temporal Dynamics:} Videos possess unique temporal fingerprints and inconsistencies resulting from the generation process, entirely missed by static image analysis. (2) \textbf{Increased Model Diversity:} The video generation pipeline involves more diverse architectures and stages (e.g., frame synthesis, motion models), creating a vastly larger and more complex attribution space. (3) \textbf{Video Compression:} Unlike image compression, video codecs introduce complex spatio-temporal artifacts that can obscure or destroy subtle generator-specific traces. These challenges necessitate a novel approach designed specifically for the video domain.

To address this significant gap, we introduce \methodname{} (\textit{\underline{S}ource \underline{A}ttribution of \underline{G}enerative \underline{A}I videos}), the first large-scale, comprehensive framework dedicated to multi-granular source attribution of AI-generated videos. Moving decisively beyond binary detection, \methodname{} pinpoints the origin of a synthetic video across five crucial levels (denoted ``\texttt{-L}") of granularity \textit{using only 0.5\% of the data}: (1) \binarylevel{} (real/synthetic); (2) \tasklevel{} (real vs. T2V vs. I2V); (3) \sdlevel{} (differentiating between Stable Diffusion versions e.g., \cite{rombach2022high, podell2023sdxl}); (4) \teamlevel{} (attributing to development teams, aiding misuse tracking); and (5) \generatorlevel{} (precise model ID). This multi-granular approach is crucial in practice: for example, when two generators are highly similar, an in-the-wild video may yield low-confidence predictions at the \generatorlevel{}, but higher confidence at coarser levels such as SD version or team, still providing valuable forensic insight. Furthermore, unlike prior works, \methodname{} provides \textit{\underline{T}emporal Attention \underline{Sig}natures} or \temporalsign{} as shown in Fig. \ref{fig:teaser}(a), which \textbf{\textit{offer crucial interpretability into why generative models are distinguishable}}.

By averaging frame-to-frame attention scores across multiple videos from a common source, we derive unique visual `fingerprints' (\temporalsign{}) for each generator. To the best of our knowledge, this is the first work to visually explain video attribution performance: \temporalsign{s} highlight the subtle but stable temporal artifacts, such as characteristic motion dynamics or frame-to-frame inconsistencies, that \methodname{} learns to leverage for fine-grained source identification.

\begin{table}[]
\centering
\caption{\textbf{Characteristic Comparison:} Unlike prior methods, \methodname{} performs video source attribution with only 0.5\% labeled data, is evaluated on a large corpus of generators from open-source datasets and provides interpretable analyses.}
\vspace{-1em}
\resizebox{0.9\linewidth}{!}{
\begin{tabular}{c|c|c}
\hline \rowcolor{\topcolor}
\textbf{Aspect}                               & \textbf{Existing Methods}  & \methodname{} \\ \hline
Binary Classification                         & {\color[HTML]{009901}{\ding{51}}} & {\color[HTML]{009901}{\ding{51}}}       \\ 
%Finegrained classification                    & \textcolor{red}{\ding{55}} & {\color[HTML]{009901}{\ding{51}}}             \\ 
Source Attribution                            & \textcolor{red}{Generator-level} & {\color[HTML]{009901}{Multi-tiered}}         \\ 
Number of Generators Evaluated                & \textcolor{red}{4 $\downarrow$}     & {\color[HTML]{009901}{20 $\uparrow$}}            \\
Data-Efficient Training                       & \textcolor{red}{\ding{55}}          & {\color[HTML]{009901}{\ding{51}}}     \\
Intra-data evaluation                         & {\color[HTML]{009901}{\ding{51}}} & {\color[HTML]{009901}{\ding{51}}}       \\ 
Cross-data evaluation                         & \textcolor{red}{\ding{55}} & {\color[HTML]{009901}{\ding{51}}}              \\ 
Quantitative Evaluation                       & {\color[HTML]{009901}{\ding{51}}} & {\color[HTML]{009901}{\ding{51}}}       \\ 
Qualitative Analysis                          & \textcolor{red}{\ding{55}} & {\color[HTML]{009901}{\ding{51}}}              \\ \hline
\end{tabular}
}
\label{tab:rel_work}
\vspace{-2em}
\end{table}

To achieve this, \methodname{} employs a novel multi-headed attention video transformer to effectively capture temporal inconsistencies. To enhance in-the-wild robustness, we initialize our model with rich visual features from a foundational vision encoder \cite{alabdulmohsin2024getting}, mitigating domain gap issues \cite{chen2025finger, lv2024domainforensics, chen2021featuretransfer}. Addressing the common scenario of abundant binary labels but scarce multi-class source labels, we propose a pretrain-and-adapt strategy. We first build a strong visual representation by pretraining a binary (real vs. fake) classifier. Subsequently, this base model is efficiently adapted to the multi-class source attribution challenge, utilizing a contrastive objective with hard-negative mining (HNM). Remarkably, this adaptation, allows \methodname{} to match the performance of a model trained with 100\% of the source-labeled data, \textit{even when using only 0.5\% of the labeled examples} for the adaptation phase (Fig. \ref{fig:teaser}(c)). This highlights exceptional data efficiency for the complex task of synthetic video source attribution.

\begin{figure*}
    \centering
    \includegraphics[width=\textwidth]{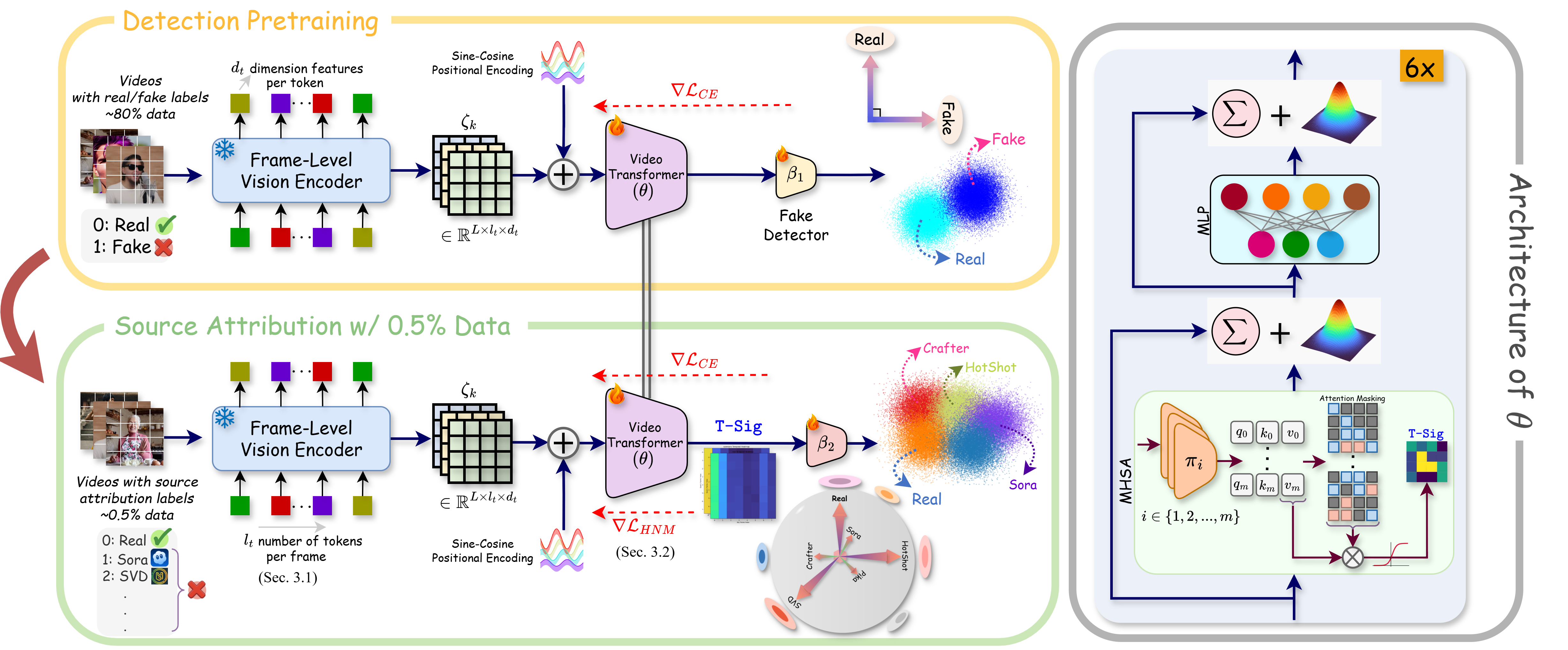}
    \vspace{-2em}
    \caption{Overall framework of \methodname{} with a two-stage training approach. In Stage-1, each video $x_k$ with real/fake labels is processed through a frozen foundational vision encoder to extract image-level features $z_m$, which are stacked in temporal order to form the video representation $\zeta_k$. Positional encoding is added, and the sequence is passed through our video transformer architecture $\theta$ (Sec. \ref{subsec:transformer}) to obtain $\phi_k$. The classifier $\beta_1$ maps $\phi_k$ to real or fake classes using a cross-entropy loss ($\mathcal{L}_{CE}$). In Stage-2, the pretrained video transformer is adapted for attribution into $n_c$ classes ($n_c$ defined by the attribution task; see supplementary Sec. S-1) using only 0.5\% of source labeled data. Stage-2 incorporates an additional hard negative mining objective ($\mathcal{L}_{\text{HNM}}$, Sec.~\ref{subsec:contrastive}) along with $\mathcal{L}_{CE}$ for the attribution task.}
    \label{fig:overall}
\vspace{-1em}
\end{figure*}

In summary, our main contributions are as follows:
\begin{itemize}
    \item We pioneer large-scale, AI-generated \textbf{video source attribution} by introducing \methodname{}. This framework moves beyond traditional binary (real/fake) detection, addressing the more complex challenge of identifying the specific origin of synthetic videos. \methodname{} demonstrates exceptional data efficiency, achieving robust attribution performance using only 0.5\% of the labeled source data per class, on par with fully supervised methods.

    \item We establish the first comprehensive, \textbf{multi-granular framework} for video source attribution, encompassing five distinct levels: \binarylevel{} (authenticity), \tasklevel{} (T2V/I2V), \sdlevel{} (base model version), \teamlevel{} (development team origin), and \generatorlevel{} (specific model). This hierarchy provides richer, more practical forensic insights than possible with single-level analysis.

    \item We introduce \textbf{Temporal Attention Signatures} (\temporalsign{s}), a novel interpretability method specifically designed for AI-generated video source attribution. Derived from \methodname{}'s learned temporal attention patterns, \temporalsign{s} provide the first visual means in the synthetic video literature to understand why different generators are distinguishable, by exposing their unique, inherent temporal artifacts.

    \item We conduct extensive evaluations across \textbf{19 distinct video generators} from two public datasets (DeMamba \cite{chen2024demamba} and DVF \cite{song2024dvfdataset}). Our results validate \methodname{}'s effectiveness and robustness in multi-granular source attribution under both in-domain and cross-domain scenarios, setting a new benchmark for this emerging field.
\end{itemize}
\section{Related Work}\label{sec:rel_work}
\noindent
\textbf{Synthetic Content Detection:} Early efforts in synthetic content detection primarily targeted images, a focus driven by the evolution of generative models, from early GAN models \cite{karras2020analyzing, karras2019style, zhu2017unpaired} to the more recent and powerful diffusion models~\cite{rombach2022high, ramesh2022hierarchical,saharia2022photorealistic}. The authors in \cite{corvi2023detection,corvi2023intriguing} investigated the detection of diffusion-generated images from GAN detectors through spatial and frequency domain analyses, and revealed the distinctive forensic traces left by generation models. Several studies explored diverse features to distinguish realistic diffusion-generated images, including reconstruction errors\cite{wang2023dire}, CLIP-based representations \cite{ojha2023towards, tan2025c2p}, and up-sampling artifacts \cite{tan2024rethinking}. However, as demonstrated in \cite{chen2024demamba, vahdati2024beyond}, image-centric approaches fall short when applied to videos, where the capture of different spatial traces or temporal artifacts is essential for effective detection. Synthetic video detection remains comparatively underexplored but has gained more focus in the recent two years. DeMamba \cite{chen2024demamba} addresses this by employing a structured state space model that continuously scans spatial and temporal zones to capture subtle generative artifacts, enabling robust real/fake classification across diverse video generators and outperforming image-based detectors on video-specific inconsistencies. A large AI-generated video dataset with 19 T2V/I2V generators is also proposed. UNITE \cite{kundu2025towards} tackles face-manipulation and synthetic video detection using a foundation model with a transformer and attention-diversity loss. While these methods advance video authenticity detection, they are primarily confined to binary classification and do not address the more challenging task of source attribution.

\noindent
\textbf{Source Model Attribution:} Research in source attribution has been largely focused on synthetic images. \cite{girish2021towards} proposed an open-world discovery and attribution pipeline that iteratively combines out-of-distribution detection, clustering, and supervised refinement, enabling the discovery and attribution of images from both known and unknown GANs in a scalable manner. POSE \cite{yang2023progressive} further advanced this by simulating open-set samples using lightweight augmentation models to better model the imperceptible traces left by unknown generative models. Wang et al. \cite{wang2023did} tackled the origin attribution problem from a model-agnostic and alteration-free perspective, proposing a reverse-engineering approach that leverages reconstruction loss: if an image can be more accurately inverted by a given model, it is likely to have been generated by that model. Collectively, these works highlight the shift from simple real/fake detection to fine-grained, open-set, and model-agnostic attribution in images. However, most of these methods are tailored for static images and cannot address the unique spatiotemporal challenges of source attribution in synthetic videos, where temporal consistency and motion artifacts play a critical role.

To the best of our knowledge, the only prior work to attempt source attribution in videos is by Vahdati et al. \cite{vahdati2024beyond}, whose study is limited to only 4 generators with closed-source videos and focuses on only generator-level attribution. In contrast, \methodname{} provides a far more comprehensive benchmark (as shown in Table \ref{tab:rel_work}) on 19 video generators, spanning multiple levels of attribution granularity along with interpretable \temporalsign{} analyses for actionable provenance analysis.
\section{Proposed Method}\label{sec:method}
Given a video $x_k$, the goal is to predict its source label $y_k$ from a set of $n_c$ possible classes. At the binary level (\binarylevel{}), we have $n_c=2$ and for source attribution, $n_c>2$. Given a dataset ${(x_k, y_k)}_{k=1}^N \in \mathcal{X}$, the \methodname{} model learns to map $x_k$ to $y_k$ under the source attribution setting, supporting both binary and fine-grained multi-class source attribution tasks.

Instead of training a $n_c$-class model from scratch, we introduce a two-stage training protocol that builds the source attribution model on top of a pre-trained binary classifier trained with extensive real/fake data. In stage-1, we pretrain a video transformer model (Sec. \ref{subsec:transformer}) for binary real vs. fake classification, as they are abundantly available, using only cross-entropy (CE) loss. In stage-2, we perform source attribution the model through a contrastive objective (Sec. \ref{subsec:contrastive}), using only 0.5\% of source labeled examples to efficiently adapt to fine-grained attribution. The pretraining is done once, and it acts as the common starting point for all levels of attribution.

\subsection{Video Transformer}\label{subsec:transformer}
AI-generated videos inherently exhibit domain gap \cite{chen2025finger, chen2021featuretransfer, lv2024domainforensics}, which is critical to address since the aim of \methodname{} is to be used in-the-wild. To enhance robustness, we extract potentially domain-agnostic features by leveraging a powerful visual encoder pretrained on web-scale image-text data. Given a video instance $x_k \in \mathcal{X}$, we process each frame $g_m$ (resized to a fixed resolution) using the frozen pretrained encoder. This produces a tokenized embedding $z_m \in \mathbb{R}^{l_t \times d_t}$ for each frame, where $m \in {1, 2, \ldots, L}$ with $L$ denoting the number of frames per video. The dimension of these embeddings is determined by the chosen encoder, where $l_t$ is the number of tokens per frame and $d_t$ is the token feature dimension. The embeddings for all frames in $x_k$ are concatenated in temporal order, resulting in a video-level representation $\zeta_k \in \mathbb{R}^{L \times l_t \times d_t}$, which serves as input to our trainable video transformer. The resulting set of encoded videos is thus represented as $\mathcal{Z} = {\zeta_k \mid x_k \in \mathcal{X} }$. The \methodname{}'s video-transformer model employs a multi-head self-attention (MHSA) transformer architecture \cite{vaswani2017attention} ($\theta$) tailored for video attribution to obtain $\phi_k=\theta(\zeta_k)$. It processes sequences of frame embeddings, effectively capturing temporal dependencies for robust video-level predictions.

Our novel video transformer architecture processes the frame-level token embeddings $\zeta_k \in \mathbb{R}^{L \times l_t \times d_t}$ in a hierarchical manner: first, by applying spatial self-attention within each frame's tokens, and second, by applying temporal self-attention across the frame-level representations.

\noindent
\textbf{Spatial Encoder:} To capture relationships between spatial patches within individual frames, the input tokens for each frame are initially processed independently. We employ a single standard transformer encoder block (detailed below). This block refines the $l_t$ token embeddings for each of the $L$ frames. The output tokens for each frame are then average pooled across the token dimension, resulting in a single feature vector $\in \mathbb{R}^{d_t}$ for each frame.

\noindent
\textbf{Temporal Encoder:} The sequence of $L$ frame-level feature vectors is then passed to the Temporal Encoder. Sinusoidal positional encodings are added to these vectors to inject temporal order information. The Temporal Encoder consists of $D = \text{depth} + 1$ stacked standard transformer encoder blocks. Each of these encoder blocks contains:

\begin{itemize}
    \item A Multi-Head Self-Attention (MHSA) layer with $N_h=12$ parallel attention heads, using scaled dot-product attention to model inter-frame dependencies.
    \item Layer Normalization, residual connections, and dropout, to ensure training stability and prevent overfitting.
    \item A two-layer feed-forward network (MLP) with GELU activation~\cite{hendrycks2016gaussian} for non-linear transformations.
\end{itemize}

This stacked architecture allows the model to build progressively complex representations of temporal dynamics and inconsistencies. The Temporal Attention Signatures (\temporalsign{s}) are extracted from the attention scores of the MHSA layer in the penultimate block encoder block of this Temporal Encoder. During inference, the attention scores over several videos are extracted and normalized to produce \temporalsign{s}. These attention scores highlight which frames the model attends to when processing the sequence, revealing patterns characteristic of the video's source.
%%%%%%%%%%%%
\subsection{Contrastive Objective}\label{subsec:contrastive}
With a pre-trained binary classifier as the foundation, Stage-2 adapts the model for multi-class source attribution. To address the limited availability of fine-grained labeled data, we incorporate a contrastive loss with hard negative mining (HNM), enabling effective attribution even with a small number of samples per generator, since CE-loss alone proved to be suboptimal in this scenario (Table \ref{tab:generator_level}, Fig. \ref{fig:tsne_generator_losses}). Given an anchor embedding $\mathbf{a}$, a positive embedding $\mathbf{p}$ (same class), and a negative embedding $\mathbf{n}$ (different class), the triplet loss encourages the following margin constraint:
\begin{equation}\small
    \|\mathbf{a} - \mathbf{p}\|_2^2 + \alpha < \|\mathbf{a} - \mathbf{n}\|_2^2,
\end{equation}
where $\alpha > 0$ is a margin hyperparameter. The loss is defined as:
\begin{equation}\small
    \mathcal{L}_{\text{triplet}} = \max\left(0, \|\mathbf{a} - \mathbf{p}\|_2^2 - \|\mathbf{a} - \mathbf{n}\|_2^2 + \alpha\right).
\end{equation}

Semi-hard negatives, which are most commonly used in the literature \cite{schroff2015facenet, harwood2017smart, xuan2020improved}, are those that are further from the anchor than the positive, but within the margin as,
\begin{equation}\label{eq:3}\small
    \|\mathbf{a} - \mathbf{p}\|_2^2 < \|\mathbf{a} - \mathbf{n}\|_2^2 < \|\mathbf{a} - \mathbf{p}\|_2^2 + \alpha.
\end{equation}
Thus, for each anchor-positive pair, the negative $\mathbf{n}$ is selected such that for a batch $\mathcal{B}$:

\begin{equation}\small
\begin{aligned}
\mathcal{L}_{\text{semi-HNM}} =& \frac{1}{|\mathcal{B}|} \sum_{i=1}^{|\mathcal{B}|} \max\Bigg(0, \|\mathbf{a}_i - \mathbf{p}_i\|_2^2 - \\
 & \min_{\substack{j \\ y_j \neq y_i \\ \|\mathbf{a}_i - \mathbf{p}_i\|_2^2 < \|\mathbf{a}_i - \mathbf{n}_j\|_2^2 \\ < \|\mathbf{a}_i - \mathbf{p}_i\|_2^2 + \alpha}} \|\mathbf{a}_i - \mathbf{n}_j\|_2^2 + \alpha \Bigg)
\end{aligned}
\end{equation}

Hard negatives are those that are closer to the anchor than the positive that is, $\|\mathbf{a} - \mathbf{n}\|_2^2 < \|\mathbf{a} - \mathbf{p}\|_2^2$. Thus, the HNM loss is:
\begin{equation}\small
    \mathcal{L}_{\text{HNM}} = \frac{1}{|\mathcal{B}|} \sum_{i=1}^{|\mathcal{B}|} \max\left(0, \|\mathbf{a}_i - \mathbf{p}_i\|_2^2 - \min_{\substack{j \\ y_j \neq y_i}} \|\mathbf{a}_i - \mathbf{n}_j\|_2^2 + \alpha\right).
\end{equation}

\begin{wrapfigure}{r}{0.4\columnwidth}
\vspace{-1em}
    \centering
    \includegraphics[width=\linewidth]{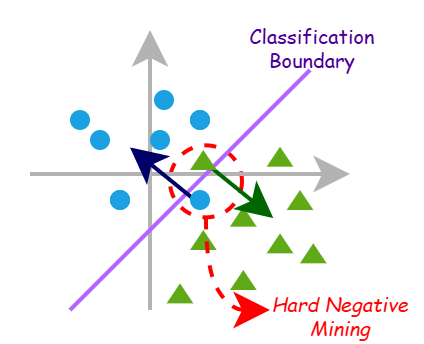}
    \vspace{-1em}
    \caption{HNM enables better separation boundaries between classes while semi-HNM will exclude these samples from the loss.}
    \label{fig:hnm}
\vspace{-1em}
\end{wrapfigure}

This focuses the model on the most challenging negatives within the batch. In our source attribution task, some generators produced embeddings with overlapping t-SNE clusters when trained with CE-loss alone (Fig. \ref{fig:tsne_generator_losses}). This is because CE-loss maximizes class separation in logit space but does not enforce geometric separation in the embedding space. Semi-hard negative mining selects negatives that satisfy  Eq. \ref{eq:3} , i.e., they are farther than the positive but still within the margin and therefore yield non‑zero loss; by contrast, easy negatives satisfy $\|\mathbf{a} - \mathbf{n}\|_2^2 > \|\mathbf{a} - \mathbf{p}\|_2^2 + \alpha$ and contribute zero loss, while hard negatives satisfy $\|\mathbf{a} - \mathbf{n}\|_2^2 \leq \|\mathbf{a} - \mathbf{p}\|_2^2$. In heavily overlapping clusters many negatives are hard rather than semi‑hard, and semi‑hard mining omits them (Fig. \ref{fig:hnm}), which limits the gradient signal needed to separate overlapping modes.

HNM, on the other hand, always selects the most difficult negative within the batch:
\begin{equation}\small
    \mathbf{n}_{\text{hard}} = \arg\min_{\substack{j \\ y_j \neq y_i}} \|\mathbf{a}_i - \mathbf{n}_j\|_2^2,
\end{equation}
and the gradient is given by:
\begin{equation}\small
    \nabla_{\theta}\mathcal{L}_{\text{HNM}} \propto 2(\mathbf{a} - \mathbf{n}_{\text{hard}}) - 2(\mathbf{a} - \mathbf{p}).
\end{equation}
This mechanism directly pushes the anchor away from the nearest negative, forcing separation even when clusters overlap. Let $\mathcal{S}_c$ denote the embedding manifold for class $c$. For overlapping classes $c$ and $c'$, hard mining minimizes,
\begin{equation}\small
    \min_{\theta} \mathbb{E}_{(\mathbf{a},\mathbf{p}) \sim \mathcal{S}_c} \left[ \max_{\mathbf{n} \sim \mathcal{S}_{c'}} \left( \|\mathbf{a} - \mathbf{p}\|_2^2 - \|\mathbf{a} - \mathbf{n}\|_2^2 + \alpha \right) \right],
\end{equation}
which is equivalent to maximizing the minimum inter-class margin $\alpha$. This is particularly important when the intra-class variance $\sigma_{\text{intra}}^2$ is comparable to or exceeds the difference between the margin and inter-class variance $\sigma_{\text{inter}}^2$ as, $\sigma_{\text{intra}}^2 \geq \alpha - \sigma_{\text{inter}}^2$. Thus, along with the CE-loss ($\mathcal{L}_{CE}$), the final loss function to train \methodname{} becomes $\lambda\cdot \mathcal{L}_{CE} + (1-\lambda)\cdot \mathcal{L}_{HNM}$.

In our experiments, CE-loss with semi-HNM was insufficient to separate overlapping generators, resulting in a mean accuracy of $70.31\%$ on the \generatorlevel{} task (see t-SNE visualizations in Fig. \ref{fig:tsne_generator_losses} and quantitative results in the supplementary). In contrast, incorporating HNM with CE-loss markedly improved performance, achieving a mean accuracy of $94.99\%$ by effectively enforcing separation between samples from different generator classes ($\mathcal{S}c$ vs. $\mathcal{S}{c'}$).
\begin{table}[]
% \vspace{-2em}
\centering
\caption{\tasklevel{} attribution performance (Accuracy) under different settings. \methodname{} performs almost perfectly, compared to the \fulldata{100\% data} setting.}
\vspace{-1em}
\resizebox{\linewidth}{!}{
\begin{tabular}{c|c|c|c|c}
\hline \rowcolor{\lightblueclr}
\textbf{Setting}                                                                                                & \textbf{Real} & \textbf{T2V} & \textbf{I2V} & \textbf{Overall} \\ \hline
\begin{tabular}[c]{@{}c@{}}Strategy: 1-stage\\Data: 0.5\%\end{tabular}                 & 99.20\%       & 97.41\%      & 66.20\%      & 82.41\%          \\ \arrayrulecolor[gray]{0.85} \hline \arrayrulecolor{black}
\begin{tabular}[c]{@{}c@{}}Strategy: 1-stage\\Data: 100\%\end{tabular}                                       & 99.97\%       & 99.93\%      & 99.97\%      & \fulldata{99.96\%}          \\ \arrayrulecolor[gray]{0.85} \hline \arrayrulecolor{black} \rowcolor{\creamclr}
\begin{tabular}[c]{@{}c@{}}Strategy: 2-stage\\Data: 0.5\%\\\oursetting{\textit{(Our Setting)}}\end{tabular} & 99.79\%       & 99.32\%      & 91.12\%      & \oursetting{98.20\%}          \\ \hline
\end{tabular}
}
\label{tab:task_level}
\vspace{-1em}
\end{table}
\begin{table}[]
% \vspace{-2em}
\centering
\caption{\sdlevel{} attribution performance under different settings. \methodname{} performs marginally better than the \fulldata{100\% data} setting.}
\vspace{-1em}
\resizebox{\linewidth}{!}{
\begin{tabular}{c|c|c|c|c|c|c}
\hline \rowcolor{\lightblueclr}
\textbf{Setting}                                                                                            & \textbf{Real} & \textbf{SD 1.4} & \textbf{SD 1.5} & \textbf{SD 2.1} & \textbf{SDXL} & \textbf{Overall} \\ \hline
\begin{tabular}[c]{@{}c@{}}Strategy: 1-stage\\Data: 0.5\%\end{tabular}                 & 99.84\%       & 0.00\%          & 0.00\%          & 99.75\%         & 99.28\%       & 59.77\%          \\ \arrayrulecolor[gray]{0.85} \hline \arrayrulecolor{black}
\begin{tabular}[c]{@{}c@{}}Strategy: 1-stage\\Data: 100\%\end{tabular}                                   & 99.99\%       & 99.90\%         & 99.99\%         & 99.80\%         & 92.09\%       & \fulldata{98.35\%}          \\ \arrayrulecolor[gray]{0.85} \hline \arrayrulecolor{black}\rowcolor{\creamclr}
\begin{tabular}[c]{@{}c@{}}Strategy: 2-stage\\Data: 0.5\%\\\oursetting{\textit{(Our Setting)}}\end{tabular} & 99.95\%       & 97.02\%         & 98.15\%         & 99.14\%         & 98.20\%       & \oursetting{98.49\%}          \\ \hline
\end{tabular}
}
\label{tab:sd_level}
\vspace{-1em}
\end{table}
\begin{table}[]
% \vspace{-2em}
\centering
\caption{Evaluation of \methodname{} on \binarylevel{} task under in-domain and various cross-domain settings. The results demonstrate the strong generalization and robustness of \methodname{} for authenticity verification, even on unseen generators.}
\vspace{-1em}
\resizebox{\linewidth}{!}{
\begin{tabular}{c|c|c|c|c}
\hline
% \rowcolor{\lightblueclr}
\textbf{Trained on}                                                                                                                  & \textbf{Tested on}                                                                                          & \textbf{Accuracy} & \textbf{Precision} & \textbf{Recall} \\ \hline
\rowcolor{\creamclr}
\begin{tabular}[c]{@{}c@{}}All DeMamba\\ generators\\ (80\% data in training)\end{tabular}                                             & \begin{tabular}[c]{@{}c@{}}All DeMamba\\ generators\\ (20\% unseen data\\ in evaluation)\end{tabular}           & 99.94\%           & 100.00\%           & 99.89\%  \\
\arrayrulecolor[gray]{0.85} \hline \arrayrulecolor{black}
\rowcolor{\lightblueclr}
\begin{tabular}[c]{@{}c@{}}DeMamba \textit{train} set\\ generators\\ (10 generators)\end{tabular}    

& \begin{tabular}[c]{@{}c@{}}DeMamba \textit{val} set\\ generators\\ (9 generators)\end{tabular}                        & 99.86\%           & 100.00\%           & 99.72\%         \\ \arrayrulecolor[gray]{0.85} \hline \arrayrulecolor{black}
\rowcolor{\creamclr}
\begin{tabular}[c]{@{}c@{}}T2V Generators\\ (12 generators)\end{tabular}                                                             & \begin{tabular}[c]{@{}c@{}}I2V Generators\\ (4 generators)\end{tabular}                                     & 99.98\%           & 99.98\%            & 99.98\%         \\ \arrayrulecolor[gray]{0.85} \hline \arrayrulecolor{black}
\rowcolor{\lightblueclr}
\begin{tabular}[c]{@{}c@{}}SD 2.1 generators\\ (6 generators)\end{tabular}                                                           & \begin{tabular}[c]{@{}c@{}}All remaining generators\\ with known\\ SD backbones\\ (5 generators)\end{tabular} & 99.94\%           & 99.90\%            & 99.96\%         \\ \arrayrulecolor[gray]{0.85} \hline \arrayrulecolor{black}
\rowcolor{\creamclr}
\begin{tabular}[c]{@{}c@{}}Generators from:\\ Alibaba Group,\\ Stability AI,\\ Tencent AI Lab, \\ and Pika AI\\ (8 Generators)\end{tabular} & \begin{tabular}[c]{@{}c@{}}All remaining generators\\ (11 generators)\end{tabular}                          & 99.16\%           & 99.98\%            & 98.41\%         \\ \hline
% \arrayrulecolor[gray]{0.85} \hline \arrayrulecolor{black}
%\begin{tabular}[c]{@{}c@{}}All DeMamba\\ generators\\ (80\% data in training)\end{tabular}                                             & \begin{tabular}[c]{@{}c@{}}All DeMamba\\ generators\\ (20\% unseen data\\ in evaluation)\end{tabular}           & 99.94\%           & 100.00\%           & 99.89\%         \\ \hline
\end{tabular}
}
\label{tab:binary_level}
\vspace{-1em}
\end{table}
\begin{table}[]
% \vspace{-2em}
\centering
\caption{SOTA and \methodname{} comparison on DeMamba \cite{chen2024demamba} for in-domain and DVF \cite{song2024dvfdataset} for cross-domain evaluation. \textbf{Best} and \underline{second-best} performances are highlighted.}
\vspace{-1em}
%\textbf{SOTA Comparison:} \binarylevel{} evaluation of \methodname{} and SOTA DeepFake detectors, with DeMamba \cite{chen2024demamba} being in-domain and DVF \cite{song2024dvfdataset} being cross-data evaluation. \textbf{Best} and \underline{second-best} performances are highlighted.}
\resizebox{\linewidth}{!}{
\begin{tabular}{c|c|c|c|c}
\hline
\rowcolor{\lightblueclr}
\textbf{Dataset}                                                  & \textbf{Method}                                                           & \textbf{Precision}         & \textbf{Recall}           & \textbf{Accuracy} \\ \hline
\multirow{12}{*}{DeMamba \cite{chen2024demamba}} & TALL \cite{xu2023tall}                                   & 87.91\%           & \underline{88.52\%}          & 88.42\%           \\  
                                                                  & F3Net \cite{qian2020thinking}                            & 88.73\%           & 81.88\%          & 86.04\%           \\  
                                                                  & NPR \cite{tan2024rethinking}                             & 82.45\%           & 84.08\%          & 83.45\%           \\  
                                                                  & STIL \cite{gu2021spatiotemporal}                         & 87.12\%           & 82.22\%          & 85.35\%           \\  
                                                                  & MINTIME-CLIP-B \cite{chen2024demamba}                    & 91.55\%           & 87.62\%          & \underline{89.98\%}     \\  
                                                                  & FTCN-CLIP-B \cite{chen2024demamba}                       & \underline{92.21\%}           & 86.18\%          & 89.67\%           \\  
                                                                  & CLIP-B-PT \cite{chen2024demamba}                         & 44.83\%           & 81.74\%          & 41.82\%           \\  
                                                                  & DeMamba-CLIP-PT \cite{chen2024demamba}                   & 79.97\%           & 78.86\%          & 79.98\%           \\  
                                                                  & XCLIP-B-PT \cite{chen2024demamba}                        & 61.29\%           & 81.93\%          & 65.83\%           \\  
                                                                  & DeMamba-XCLIP-PT \cite{chen2024demamba}                  & 76.38\%           & 83.59\%          & 79.31\%           \\  
                                                                  & XCLIP-B-FT \cite{chen2024demamba}                        & 86.77\%           & 84.41\%          & 86.07\%           \\ \arrayrulecolor[gray]{0.85} \cline{2-5} \arrayrulecolor{black} 
                                                                  & \cellcolor{\bottomcolor}\methodname{} (\binarylevel{}) & \cellcolor{\bottomcolor}\textbf{100.00\%} & \cellcolor{\bottomcolor}\textbf{99.89\%} & \cellcolor{\bottomcolor}\textbf{99.94\%}  \\ \hline
\multirow{12}{*}{DVF \cite{song2024dvfdataset}}    & CNNDet \cite{wang2020cnn}                                & -                 & -                & 78.20\%           \\  
                                                                  & DIRE \cite{wang2023dire}                                 & -                 & -                & 62.10\%           \\  
                                                                  & Raising \cite{cozzolino2024raising}                      & -                 & -                & 67.00\%           \\  
                                                                  & UNI-FD \cite{ojha2023towards}                            & -                 & -                & 74.10\%           \\  
                                                                  & F3Net \cite{qian2020thinking}                            & -                 & -                & 81.30\%           \\  
                                                                  & ViViT \cite{arnab2021vivit}                              & -                 & -                & 79.10\%           \\  
                                                                  & TALL \cite{xu2023tall}                                   & -                 & -                & 69.50\%           \\  
                                                                  & TS2-Net \cite{liu2022ts2}                                & -                 & -                & 72.10\%           \\  
                                                                  & DE-FAKE \cite{sha2023fake}                               & -                 & -                & 72.10\%           \\  
                                                                  & HifiNet \cite{guo2023hierarchical}                       & -                 & -                & 84.30\%           \\  
                                                                  & DVF \cite{song2024dvfdataset}                            & -                 & -                & \underline{92.00\%}     \\ \arrayrulecolor[gray]{0.85} \cline{2-5} \arrayrulecolor{black}
                                                                  & \cellcolor{\bottomcolor}\methodname{} (\binarylevel{}) & \cellcolor{\bottomcolor}\textbf{99.35\%}  & \cellcolor{\bottomcolor}\textbf{96.14\%} & \cellcolor{\bottomcolor}\textbf{95.39\%}  \\ \hline
\end{tabular}
}
\label{tab:sota_comparison}
\vspace{-1em}
\end{table}
% \begin{table}[]
\begin{table}[]
% \vspace{-2em}
\centering
\caption{\teamlevel{} performances (Accuracy) under different settings. \methodname{} performs better than the \fulldata{100\% data} setting on average. The proposed 2-stage training significantly improves performance on certain teams as highlighted.}
\vspace{-1em}
% \resizebox{0.5\textwidth}{!}{
\resizebox{\linewidth}{!}{
\begin{tabular}{c|c|c|c}
\hline\rowcolor{\lightblueclr}
\textbf{Team}      & \begin{tabular}[c]{@{}c@{}}Strategy: 1-stage\\Data: 0.5\%\end{tabular} & \begin{tabular}[c]{@{}c@{}}Strategy: 1-stage\\Data: 100\%\end{tabular} & \begin{tabular}[c]{@{}c@{}}Strategy: 2-stage\\Data: 0.5\%\\\oursetting{\textit{(Our Setting)}}\end{tabular} \\ \hline
\textit{Real}      & 99.54\%                                                                                         & 99.95\%                                                                   & 98.86\%                                                                                                         \\
Alibaba Group      & 97.57\%                                                                                         & 99.80\%                                                                   & 97.63\%                                                                                                         \\
Hotshot Co.        & 94.96\%                                                                                         & 95.68\%                                                                   & 98.56\%                                                                                                         \\ 
HPC AI Tech        & 92.99\%                                                                                         & 99.95\%                                                                   & 96.85\%                                                                                                         \\ 
MoonValley         & 98.52\%                                                                                         & 100.00\%                                                                  & 100.00\%                                                                                                        \\ 
MorphStudio        & 88.27\%                                                                                         & 77.78\%                                                                   & 83.95\%                                                                                                         \\ 
OpenAI             & \cellcolor{red!35}13.33\%                                                                                         & 80.00\%                                                                   & \cellcolor{green!35}66.67\%                                                                                                         \\ 
Personal: Sterling & 60.96\%                                                                                         & 81.96\%                                                                   & 95.80\%                                                                                                         \\ 
Pika               & 88.81\%                                                                                         & 99.76\%                                                                   & 95.84\%                                                                                                         \\ 
Runway ML          & 74.62\%                                                                                         & 90.53\%                                                                   & 92.42\%                                                                                                         \\ 
Shanghai AI Lab-1  & 95.98\%                                                                                         & 99.91\%                                                                   & 97.93\%                                                                                                         \\ 
Shanghai AI Lab-2  & \cellcolor{red!35}0.00\%                                                                                          & 99.99\%                                                                   & \cellcolor{green!35}96.26\%                                                                                                         \\ 
Show Lab           & 98.10\%                                                                                         & 100.00\%                                                                  & 98.10\%                                                                                                         \\ 
Stability AI       & 91.20\%                                                                                         & 99.98\%                                                                   & 98.65\%                                                                                                         \\ 
Tencent AI Lab     & 84.24\%                                                                                         & 98.76\%                                                                   & 93.19\%                                                                                                         \\ \arrayrulecolor[gray]{0.85} \hline \arrayrulecolor{black} \rowcolor{\creamclr}
\textbf{Overall}   & 80.55\%                                                                                         & \fulldata{94.94\%}                                                                   & \oursetting{97.77\%}                                                                                                         \\ \hline
\end{tabular}
}
\label{tab:team_level}
\vspace{-2em}
\end{table}
% \end{table}
\begin{table*}[]
\centering
\caption{\generatorlevel{} classification results (Accuracy) with different settings of the \methodname{} framework. \methodname{} is able to achieve results close to the \fulldata{100\%} setting, by only using 0.5\% of source labeled data (100\% data setting has $\sim$1.6M training data). In many cases (as highlighted) the performance is close to $0.00\%$ for certain difficult generators, but the $\mathcal{L}_{HNM}$ objective has been able to mitigate these missed detections even while using a small fraction of the data. especially while using the proposed 2-stage training.} 
\vspace{-1em}
\resizebox{0.85\textwidth}{!}{
\begin{tabular}{c|cc|cc|cc}
\hline
\rowcolor{\lightblueclr}
\multirow{2}{*}{\textbf{Generators}} & \multicolumn{2}{c|}{\textbf{\begin{tabular}[c]{@{}c@{}}Strategy: 1-stage\\Data: 0.5\%\end{tabular}}}    & \multicolumn{2}{c|}{\textbf{\begin{tabular}[c]{@{}c@{}}Strategy: 2-stage\\Data: 0.5\%\end{tabular}}} & \multicolumn{2}{c}{\textbf{\begin{tabular}[c]{@{}c@{}}Strategy: 1-stage\\Data: 100\%\end{tabular}}}                        \\ \cline{2-7} \rowcolor{\lightblueclr}
                                     & \multicolumn{1}{c|}{$\mathcal{L}_{CE}$ Only} & $\mathcal{L}_{CE}+\mathcal{L}_{HNM}$ & \multicolumn{1}{c|}{$\mathcal{L}_{CE}$ Only}                            & $\mathcal{L}_{CE}+\mathcal{L}_{HNM}$ \oursetting{\textit{(Our Setting)}}     & \multicolumn{1}{c|}{$\mathcal{L}_{CE}$ Only} & $\mathcal{L}_{CE}+\mathcal{L}_{HNM}$ \\ \hline
\textit{Real}                          & \multicolumn{1}{c|}{0.00\%}                  & 99.04\%                               & \multicolumn{1}{c|}{98.12\%}                                            & 99.95\%                                                         & \multicolumn{1}{c|}{99.91\%}                 & 99.21\%                               \\ 
DynamiCrafter                        & \multicolumn{1}{c|}{0.00\%}                  & 0.00\%                                & \multicolumn{1}{c|}{0.20\%}                                             & 56.64\%                                                         & \multicolumn{1}{c|}{99.53\%}                 & 75.58\%                               \\ 
I2VGen-XL                            & \multicolumn{1}{c|}{\cellcolor{red!35}0.24\%}                  & \cellcolor{red!35}0.00\%                                & \multicolumn{1}{c|}{\cellcolor{red!35}9.19\%}                                             & \cellcolor{green!35}96.87\%                                                         & \multicolumn{1}{c|}{95.13\%}                 & 96.84\%                               \\ 
Latte                                & \multicolumn{1}{c|}{0.00\%}                  & 0.00\%                                & \multicolumn{1}{c|}{36.89\%}                                            & 98.33\%                                                         & \multicolumn{1}{c|}{99.30\%}                 & 97.84\%                               \\ 
OpenSora                             & \multicolumn{1}{c|}{98.60\%}                 & 0.00\%                                & \multicolumn{1}{c|}{1.23\%}                                             & 91.00\%                                                         & \multicolumn{1}{c|}{99.82\%}                 & 96.70\%                               \\ 
SD                                   & \multicolumn{1}{c|}{99.82\%}                 & 97.10\%                               & \multicolumn{1}{c|}{3.36\%}                                             & 91.28\%                                                         & \multicolumn{1}{c|}{98.86\%}                 & 97.76\%                               \\ 
SEINE                                & \multicolumn{1}{c|}{0.00\%}                  & 95.47\%                               & \multicolumn{1}{c|}{1.48\%}                                             & 89.25\%                                                         & \multicolumn{1}{c|}{70.79\%}                 & 95.95\%                               \\ 
SVD                                  & \multicolumn{1}{c|}{\cellcolor{red!35}0.00\%}                  & \cellcolor{red!35}0.00\%                                & \multicolumn{1}{c|}{\cellcolor{red!35}0.06\%}                                             & \cellcolor{green!35}91.45\%                                                         & \multicolumn{1}{c|}{99.63\%}                 & 96.68\%                               \\ 
VideoCrafter                         & \multicolumn{1}{c|}{0.00\%}                  & 91.25\%                               & \multicolumn{1}{c|}{27.23\%}                                            & 92.27\%                                                         & \multicolumn{1}{c|}{62.71\%}                 & 95.23\%                               \\ 
ZeroScope                            & \multicolumn{1}{c|}{0.00\%}                  & 0.00\%                                & \multicolumn{1}{c|}{68.11\%}                                            & 91.50\%                                                         & \multicolumn{1}{c|}{99.91\%}                 & 95.95\%                               \\ 
Pika                                 & \multicolumn{1}{c|}{99.58\%}                 & 93.54\%                               & \multicolumn{1}{c|}{15.01\%}                                            & 89.99\%                                                         & \multicolumn{1}{c|}{98.44\%}                 & 93.43\%                               \\ 
Crafter                              & \multicolumn{1}{c|}{0.00\%}                  & 0.00\%                                & \multicolumn{1}{c|}{78.79\%}                                            & 79.80\%                                                         & \multicolumn{1}{c|}{85.52\%}                 & 94.95\%                               \\ 
Gen2                                 & \multicolumn{1}{c|}{\cellcolor{red!35}0.38\%}                  & \cellcolor{red!35}0.00\%                                & \multicolumn{1}{c|}{69.32\%}                                            & \cellcolor{green!35}84.85\%                                                         & \multicolumn{1}{c|}{\cellcolor{red!35}11.36\%}                 & 78.79\%                               \\ 
HotShot                              & \multicolumn{1}{c|}{0.00\%}                  & 92.09\%                               & \multicolumn{1}{c|}{94.96\%}                                            & 96.40\%                                                         & \multicolumn{1}{c|}{42.45\%}                 & 95.68\%                               \\ 
Lavie                                & \multicolumn{1}{c|}{\cellcolor{red!35}0.00\%}                  & 55.94\%                               & \multicolumn{1}{c|}{79.37\%}                                            & \cellcolor{green!35}81.82\%                                                         & \multicolumn{1}{c|}{\cellcolor{red!35}3.85\%}                  & 92.66\%                               \\ 
ModelScope                           & \multicolumn{1}{c|}{0.00\%}                  & 0.00\%                                & \multicolumn{1}{c|}{87.59\%}                                            & 97.81\%                                                         & \multicolumn{1}{c|}{8.76\%}                  & 97.08\%                               \\
MoonValley                           & \multicolumn{1}{c|}{0.00\%}                  & 97.78\%                               & \multicolumn{1}{c|}{97.78\%}                                            & 99.26\%                                                         & \multicolumn{1}{c|}{97.78\%}                 & 100.00\%                              \\
MorphStudio                          & \multicolumn{1}{c|}{0.00\%}                  & 14.81\%                               & \multicolumn{1}{c|}{90.12\%}                                            & 81.48\%                                                         & \multicolumn{1}{c|}{0.00\%}                  & 82.10\%                               \\
Show\_1                              & \multicolumn{1}{c|}{0.00\%}                  & 0.00\%                                & \multicolumn{1}{c|}{92.38\%}                                            & 98.10\%                                                         & \multicolumn{1}{c|}{45.71\%}                 & 100.00\%                              \\
Sora                                 & \multicolumn{1}{c|}{0.00\%}                  & 0.00\%                                & \multicolumn{1}{c|}{93.33\%}                                            & 73.33\%                                                         & \multicolumn{1}{c|}{66.67\%}                 & 60.00\%                               \\ \arrayrulecolor[gray]{0.85} \hline \arrayrulecolor{black} \rowcolor{\creamclr}
\textbf{Overall}                     & \multicolumn{1}{c|}{24.55\%}                 & 65.80\%                               & \multicolumn{1}{c|}{55.13\%}                                            & \oursetting{94.99\%}                                                   & \multicolumn{1}{c|}{70.51\%}                 & \fulldata{97.41\%}                      \\ \hline
\end{tabular}
}
\vspace{-1em}
\label{tab:generator_level}
\end{table*}

\section{Experiments}\label{sec:experiments}
\textbf{\underline{Datasets}:} The training is performed on the DeMamba \cite{chen2024demamba} dataset, with 19 different AI video generators and 1M real videos. More details can be found in the supplementary material,
%appendix \ref{appendix:attribution_details}, 
including how we define different attribution levels. Additionally, we use the DVF dataset \cite{song2024dvfdataset} covering 8 video generators for cross-data evaluations. Implementation details are provided in the supplementary.

\noindent
\textbf{\underline{Real/Fake Detection Results}:} We first comprehensively evaluate the performance of \methodname{} in binary classification in Table~\ref{tab:binary_level}, conducting both in-domain and cross-generator evaluations. For cross-generator analysis, we train on the \textit{train} split of DeMamba and evaluate on the \textit{val} split, as well as train on generators from a specific team or SD version backbone or generation task and test on the remaining generators. The results demonstrate that, for the \binarylevel{} task, \methodname{} achieves robust authenticity verification across diverse and previously unseen data sources, exhibiting strong generalization and minimal sensitivity to domain shifts. To rigorously benchmark \methodname{} against existing state-of-the-art (SOTA) binary detectors, we conducted evaluations on both the DeMamba~\cite{chen2024demamba} and DVF~\cite{song2024dvfdataset} datasets, as presented in Table~\ref{tab:sota_comparison}. Notably, all competing SOTA methods on the DVF dataset \cite{song2024dvfdataset} are trained and tested within the DVF dataset, following standard in-domain protocols. In contrast, \methodname{} is trained exclusively on DeMamba and evaluated directly on DVF, constituting a challenging cross-dataset generalization scenario. Despite this, \methodname{} significantly outperforms the SOTA baselines, highlighting its superior robustness and generalizability for the \binarylevel{} task across diverse data distributions.

\begin{figure*}
    \centering
    \subfloat[\methodname{} (\tasklevel{}) on\\different tasks]{\includegraphics[width=0.25\textwidth]{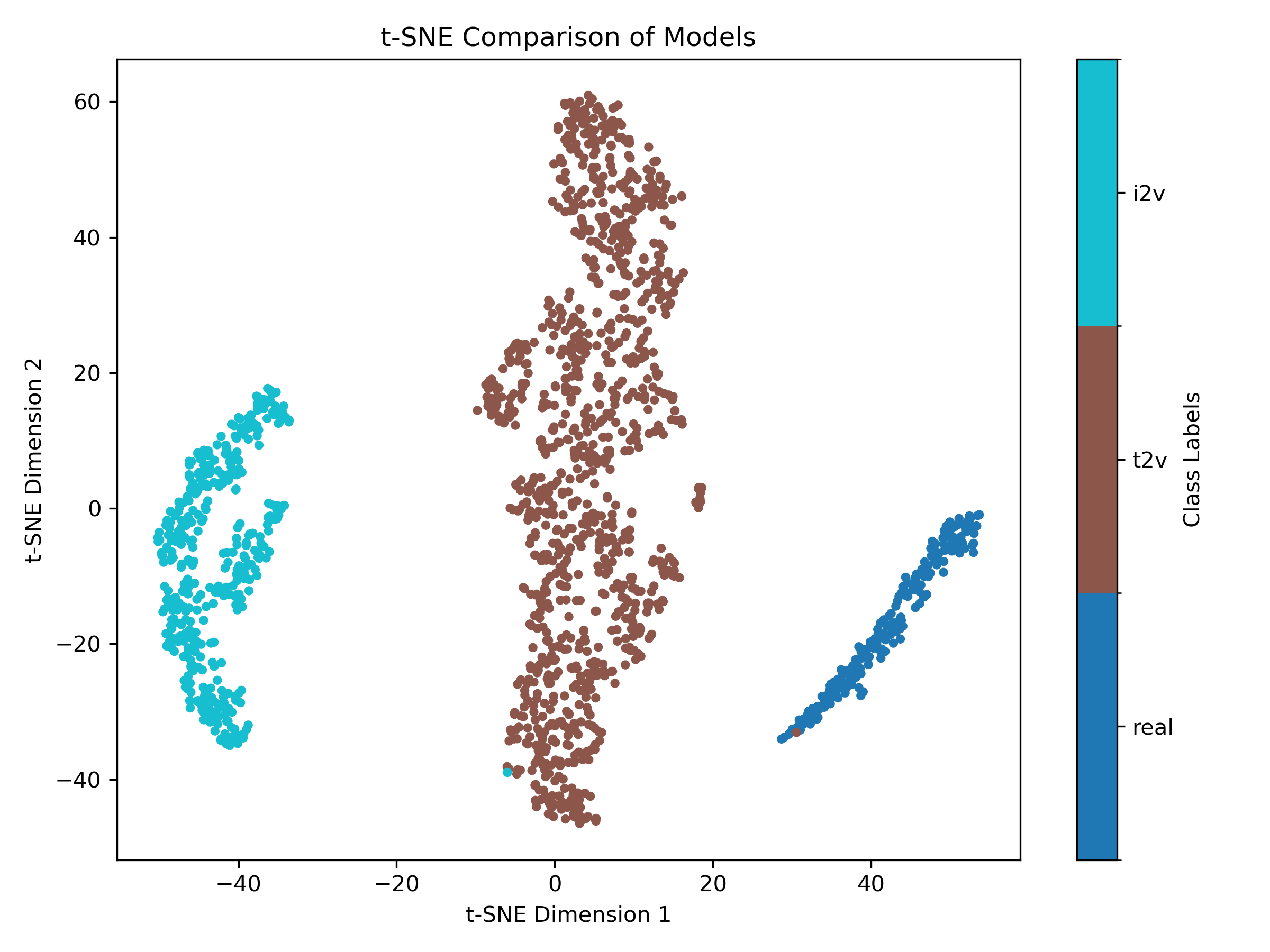}}
    \subfloat[\methodname{} (\tasklevel{}) on\\different generators]{\includegraphics[width=0.25\textwidth]{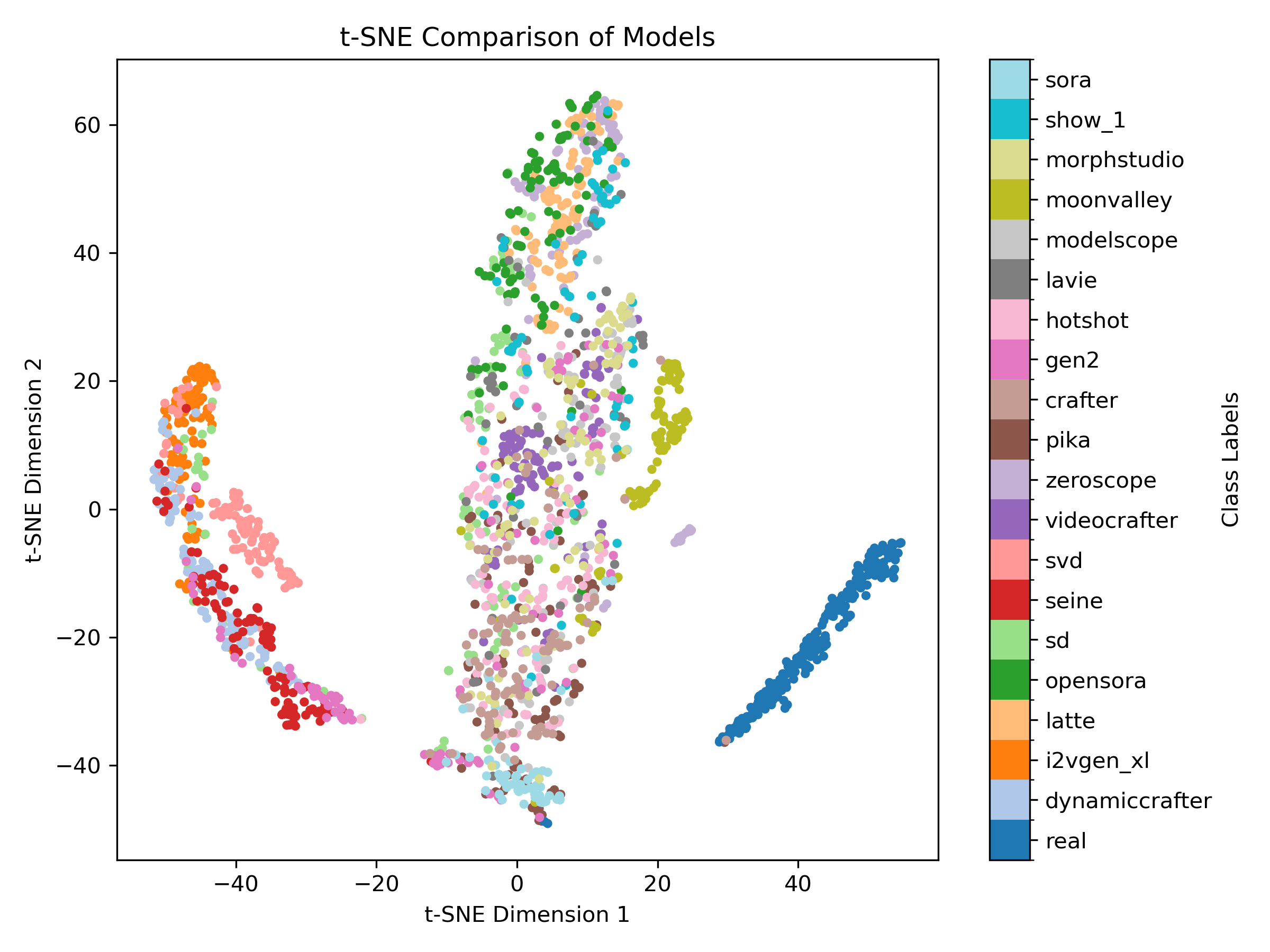}}
    \subfloat[\methodname{} (\binarylevel{}) on\\different generators]{\includegraphics[width=0.25\textwidth]{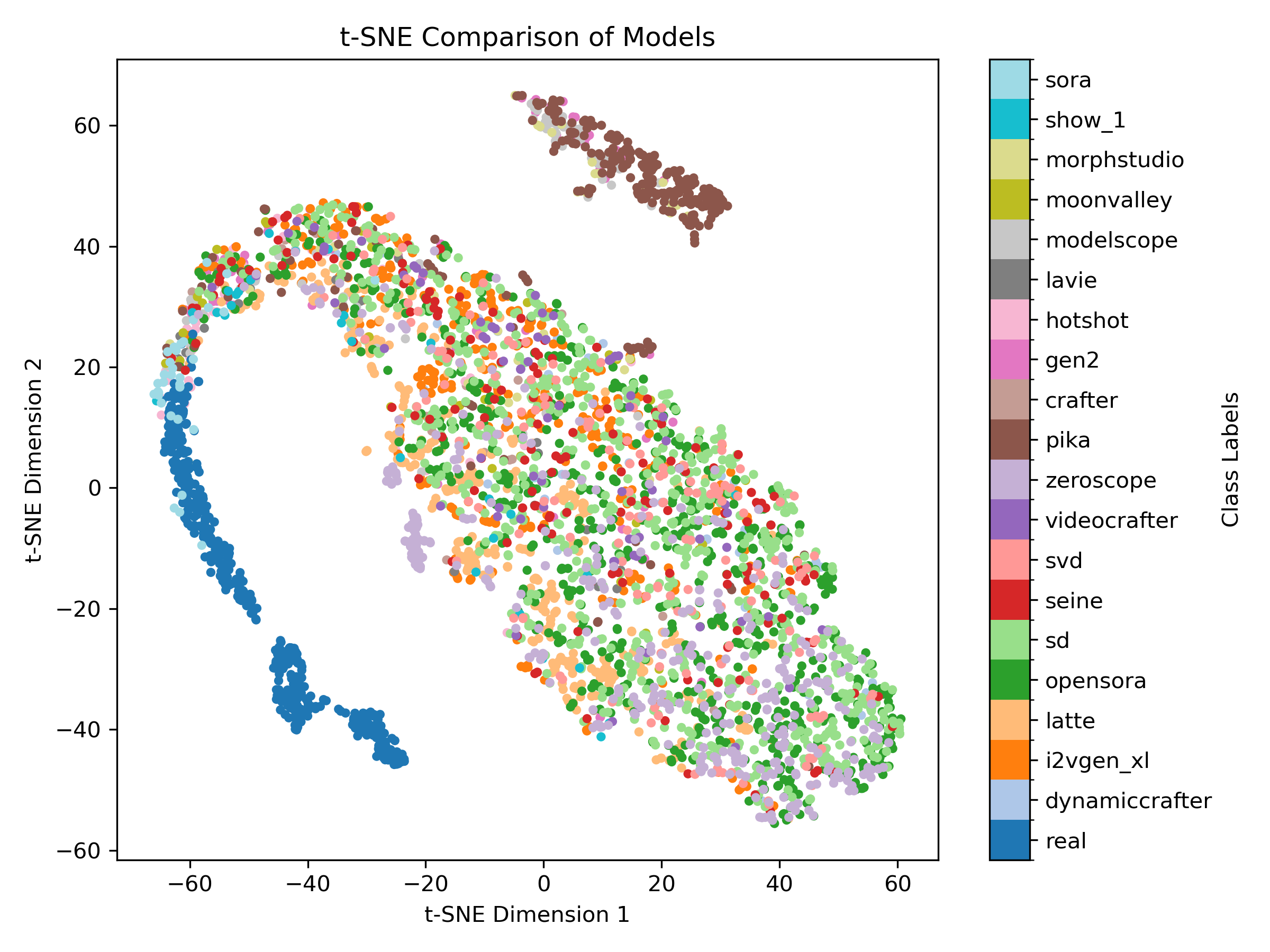}}
    \subfloat[\methodname{} (10-class) on\\all generators]{\includegraphics[width=0.25\textwidth]{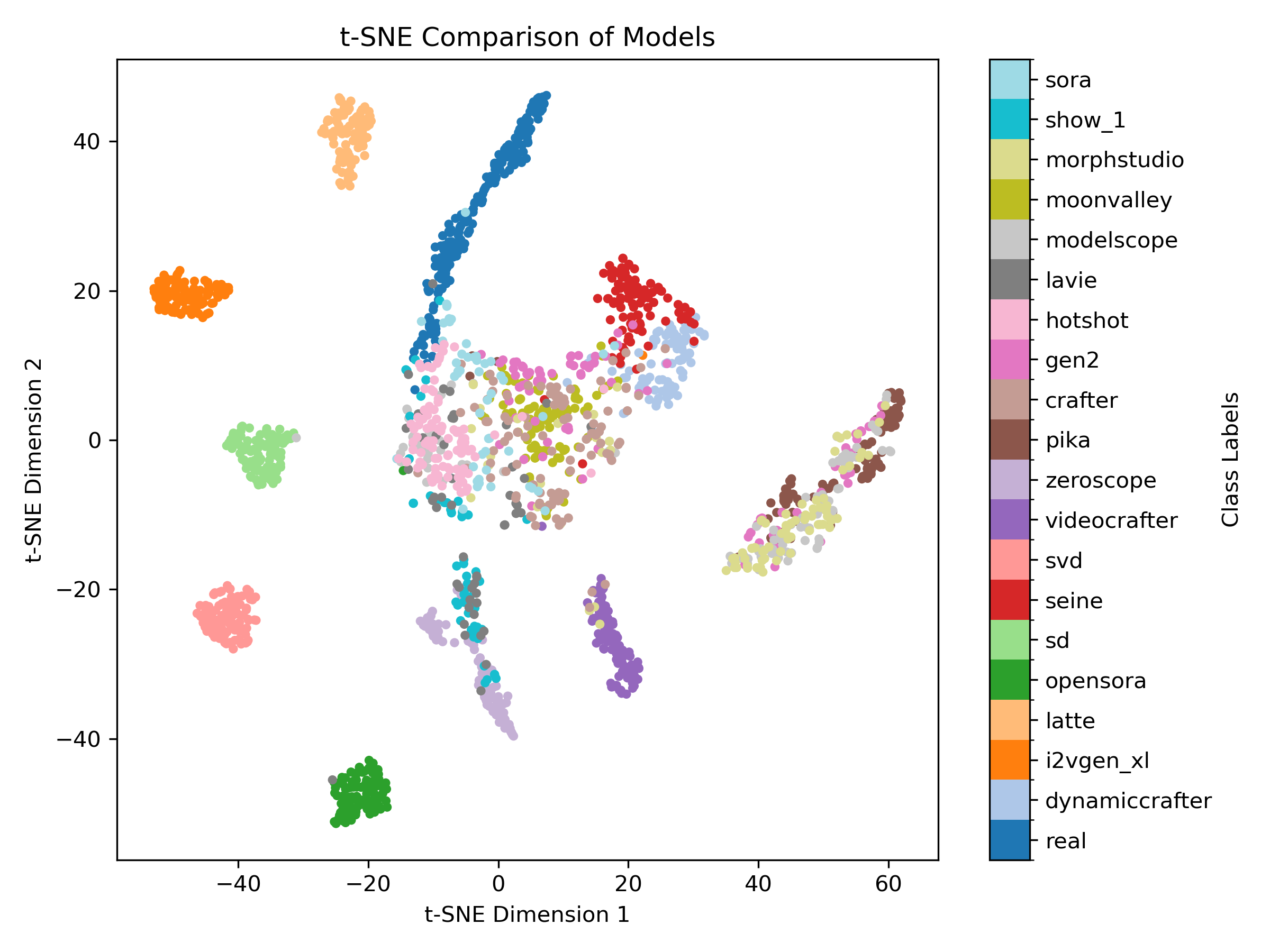}}

    \centering
    \subfloat[\methodname{} (\sdlevel{}) on\\different SD versions]{\includegraphics[width=0.25\textwidth]{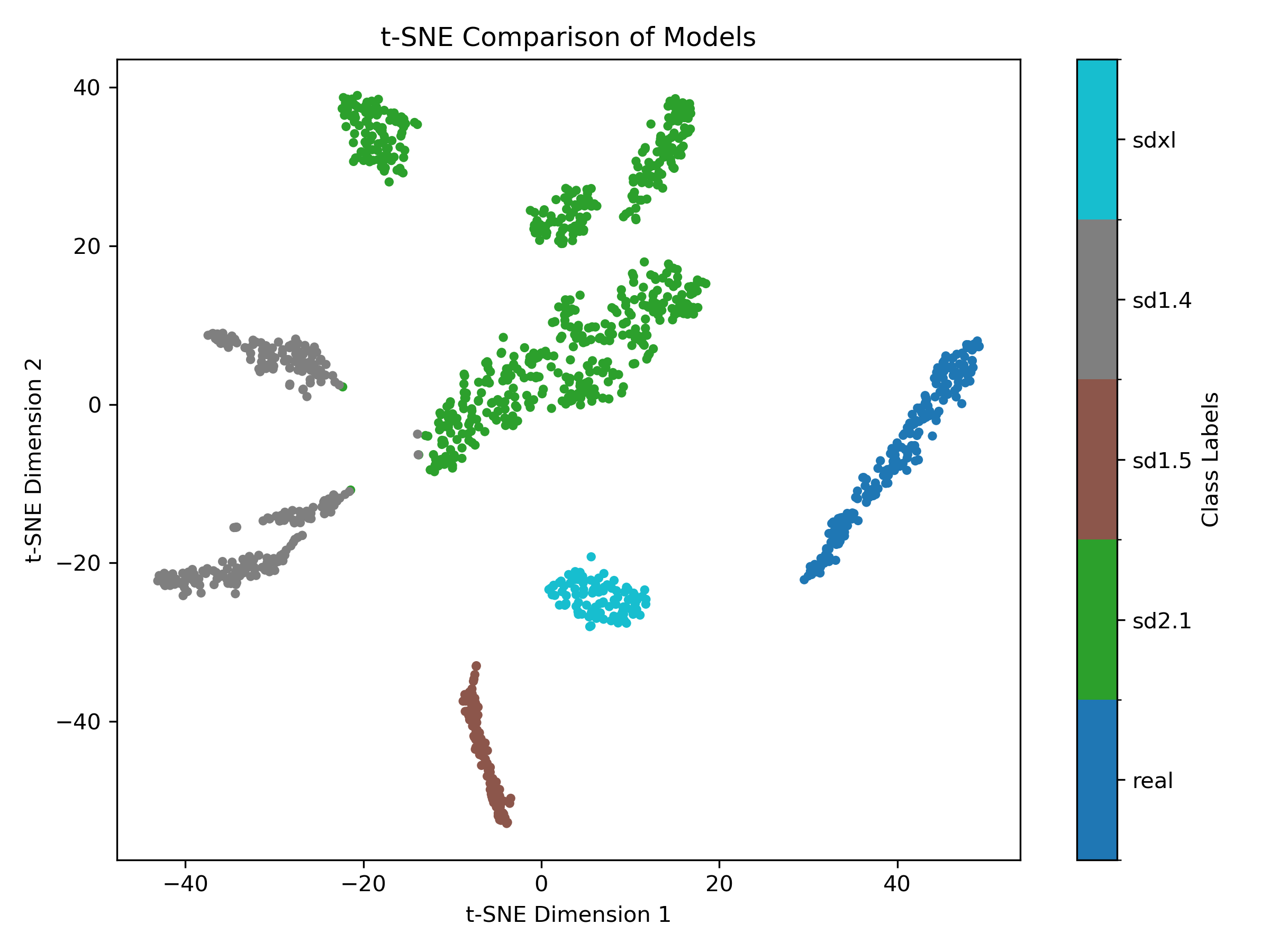}}
    \subfloat[\methodname{} (\sdlevel{}) on\\different generators]{\includegraphics[width=0.25\textwidth]{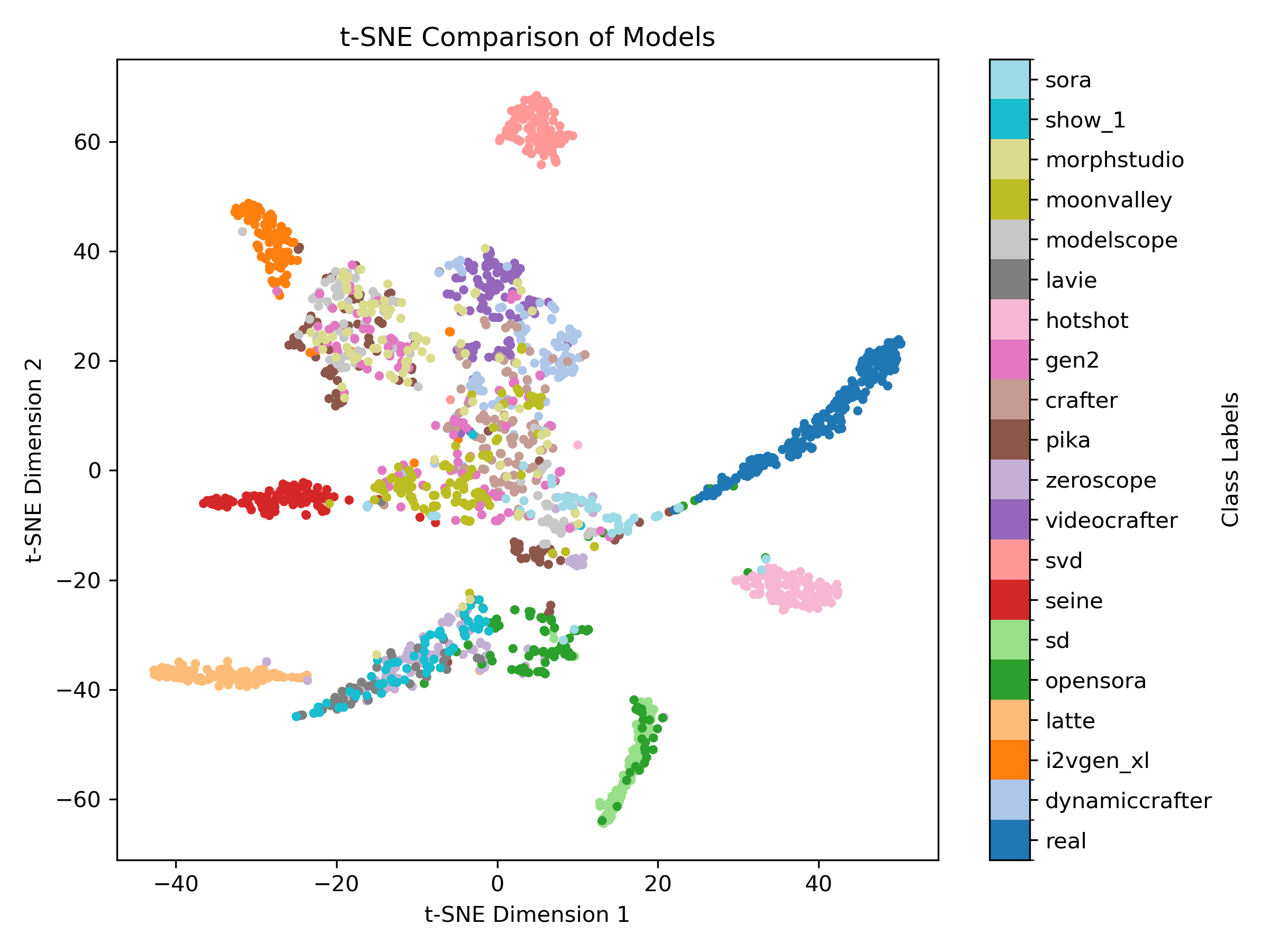}}
    \subfloat[\methodname{} (\teamlevel{}) on\\different teams]{\includegraphics[width=0.25\textwidth]{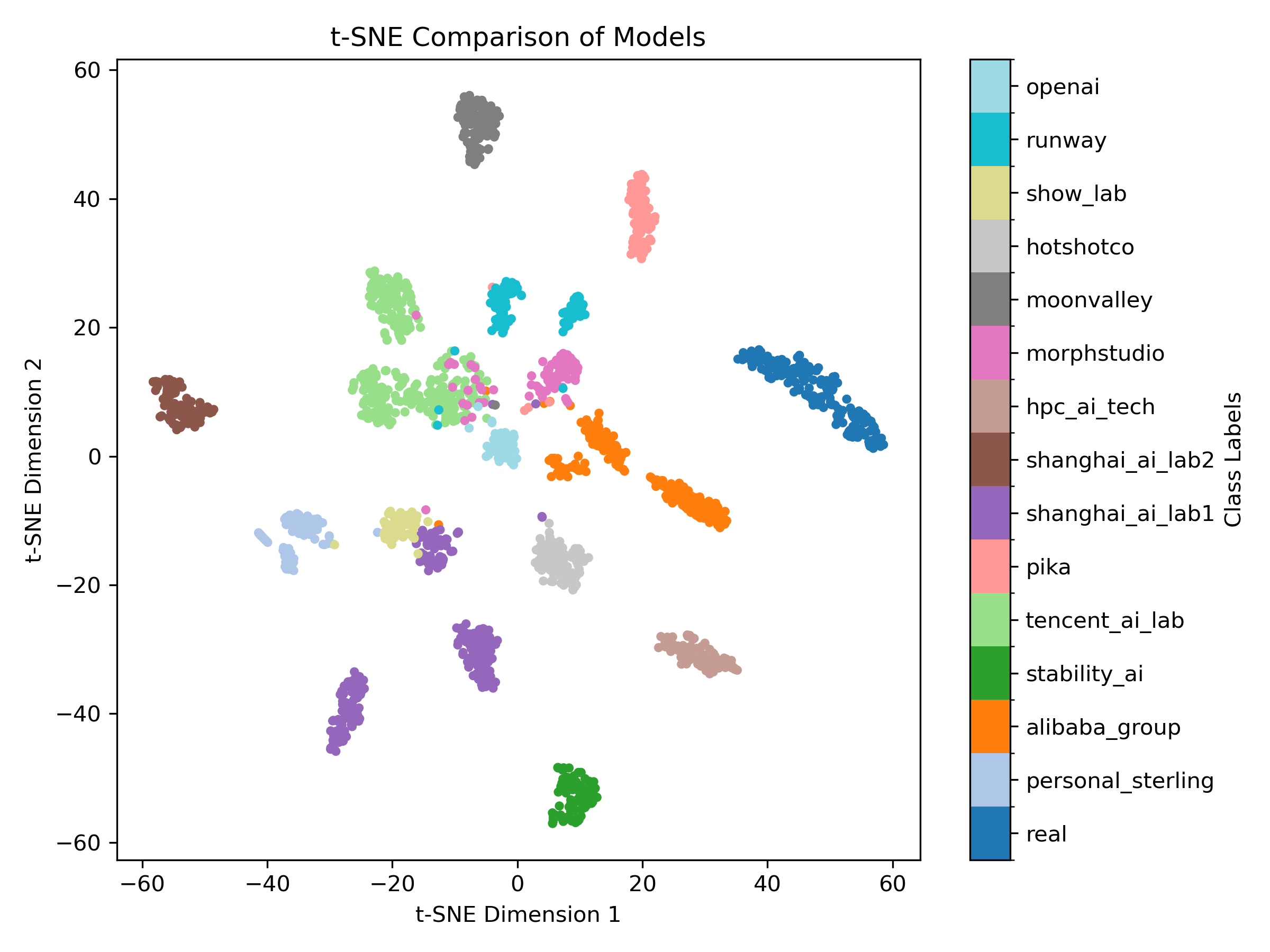}}
    \subfloat[\methodname{} (\teamlevel{}) on\\different generators]{\includegraphics[width=0.25\textwidth]{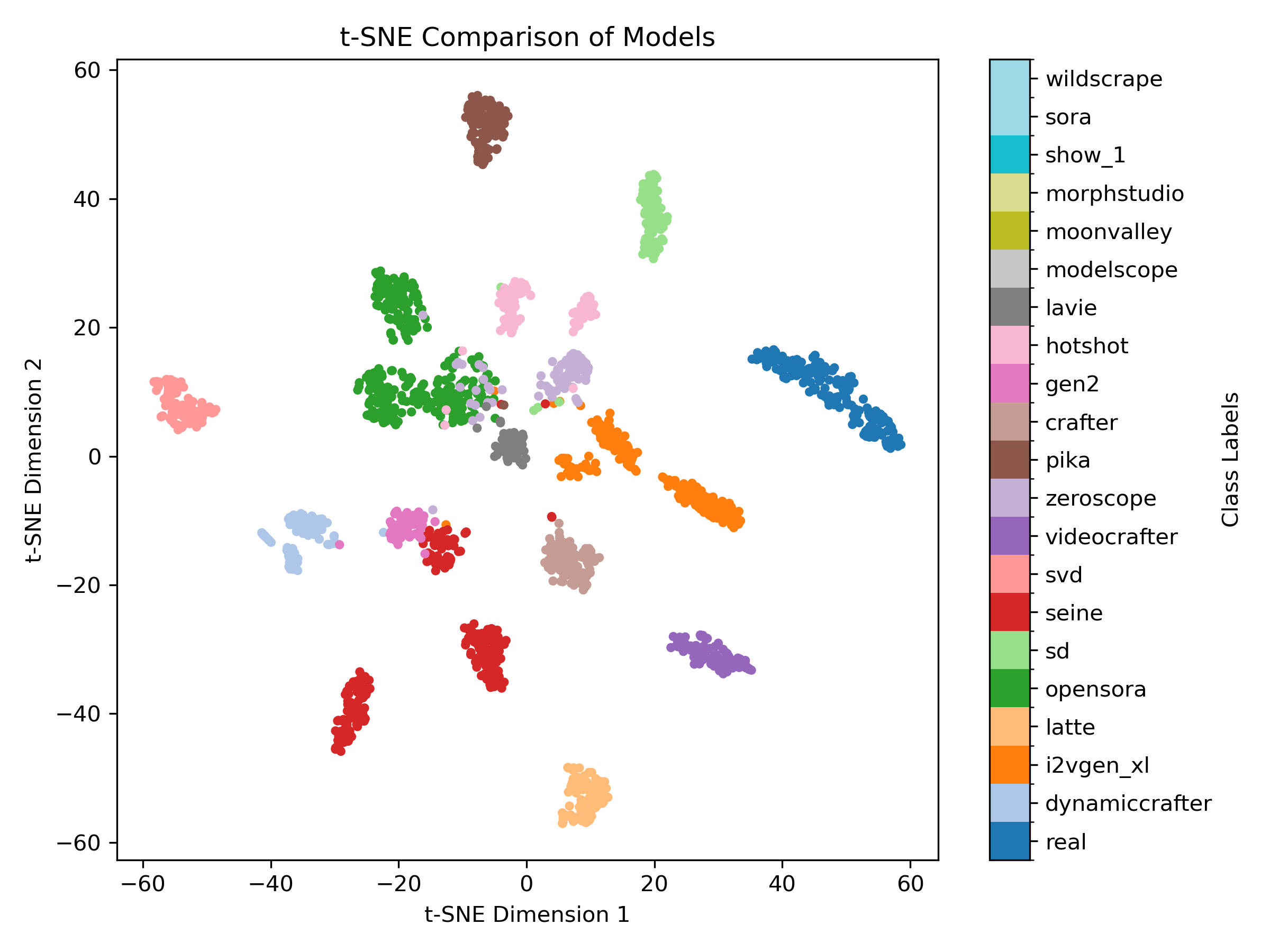}}
    \vspace{-1em}
    \caption{t-SNE visualization of \methodname{}'s learned representations trained on the \tasklevel{}, \binarylevel{}, \sdlevel{} and \teamlevel{} attribution tasks, respectively. Even when supervised at coarser levels, \methodname{} distinctly clusters individual generators, revealing strong fine-grained discriminative ability.}
    \vspace{-1em}
\label{fig:tsne_analysis}
\end{figure*}

\noindent
\textbf{\underline{Fine-grained Attribution Results}:} We further evaluate \methodname{} under multi-granular source attribution tasks across three training regimes: (1) 1-stage training with only 0.5\% labeled data, (2) 1-stage training with 100\% data, and (3) our proposed 2-stage training framework using 0.5\% labeled data for all attribution levels (attribution description in the supplementary).

On \tasklevel{} task (Table~\ref{tab:task_level}), \methodname{} achieves strong performance in distinguishing real, T2V, and I2V videos. This exposes the distinctive patterns left by T2V and I2V generation methods. With full data, the model nearly saturates accuracy across all classes (99.96\% overall). In the low-data setting, performance drops substantially for I2V (66.20\%), indicating limited data hinders generalization to this class. Our two-stage training framework substantially mitigates this drop, boosting I2V accuracy to 91.12\% and overall accuracy to 98.20\%.

Evaluation on the \sdlevel{} task in Table~\ref{tab:sd_level} demonstrates that full-data training yields high accuracy across all SD versions (98.35\% overall). However, in the low-data regime, the model struggles to distinguish SD 1.4 and SD 1.5 (both 0\%), while maintaining high accuracy for real, SD 2.1, and SDXL. The two-stage approach closes this gap, achieving over 97\% for all SD versions. % and 98.49\% overall.
On the \teamlevel{} task (Table~\ref{tab:team_level}), the model achieves high accuracy for most teams with full data (94.94\% overall). In the low-data setting, performance varies widely across teams, with some (e.g., Shanghai AI Lab-2, OpenAI) at or near 0\%. The two-stage training strategy dramatically improves robustness, yielding 97.77\% overall and consistently high accuracy across almost all teams.

Table~\ref{tab:generator_level} presents results for the most challenging setting: \generatorlevel{} attribution. Using only cross-entropy loss in the low-data regime, the model performs poorly (24.55\% overall), but adding a hard negative contrastive loss boosts accuracy to 65.80\%. Our two-stage framework with HNM achieves 94.99\% overall, a substantial improvement over single-stage approaches. With full data, the model achieves up to 97.41\% accuracy, highlighting the benefit of both data scale and contrastive learning for fine-grained attribution.

\noindent
\textbf{\underline{t-SNE Analysis}:} To further interpret the representations learned by \methodname{}, we conduct a t-SNE~\cite{van2008visualizing} and \temporalsign{} analyses of the feature embeddings produced by the model under different attribution settings. Specifically, we visualize the embedding outputs of the last ($6^{th}$) encoder of $\theta$ for videos from the validation set when the model is trained for all attribution levels. First, Fig.~\ref{fig:tsne_analysis} (a) and (b) with \tasklevel{} attribution results show that embeddings for real, T2V, and I2V samples form clearly separable clusters, demonstrating effective discrimination among these broad categories. However, when the same embeddings are colored by generator, substantial overlap is observed among most generators, with only a few, such as MorphStudio~\cite{morphstudio} and SVD~\cite{blattmann2023stable}, forming distinct clusters. This suggests that while \methodname{} is highly effective at coarse-grained attribution, it does not inherently separate individual generators at this level. Fig.~\ref{fig:tsne_analysis} (c) shows the t-SNE plot for the \methodname{} model trained for \binarylevel{} classification. Here, all generators except Pika~\cite{pika2022} collapse into a single ``fake" cluster, indicating that the model learns to aggregate all synthetic sources together for the binary task, with minimal separation among generators.

Fig. \ref{fig:tsne_analysis} (d) visualizes embeddings from a \methodname{} model trained on 10 specific generators from the DeMamba \textit{train} set. The seen generators form distinct clusters, and notably, several unseen generators such as Hotshot \cite{hotshot2023}, Show\_1 \cite{zhang2024show}, and MorphStudio \cite{morphstudio} also appear as separable clusters. This indicates that the model, even trained on a subset of generators, can recognize distributional differences and cluster unseen sources, highlighting its potential for generalization in open-set scenarios. For the \sdlevel{} and \teamlevel{} models, t-SNE projections (Fig. \ref{fig:tsne_analysis} (e) - (h)) reveal that the learned representations not only cluster according to the supervised SD version or team labels, but also often separate individual generators within each group. This indicates that \methodname{} captures fine-grained differences between generators, even when supervision is provided only at a coarser level. This level of separation indicates that the model is sensitive to subtle distributional differences introduced by specific generator architectures or research teams, enabling it to infer whether an unknown generator shares an SD backbone or team affiliation, or represents a completely novel source.

\begin{figure}
% \vspace{-1em}
    \centering
    \subfloat[CE-loss only]{\includegraphics[width=0.33\columnwidth]{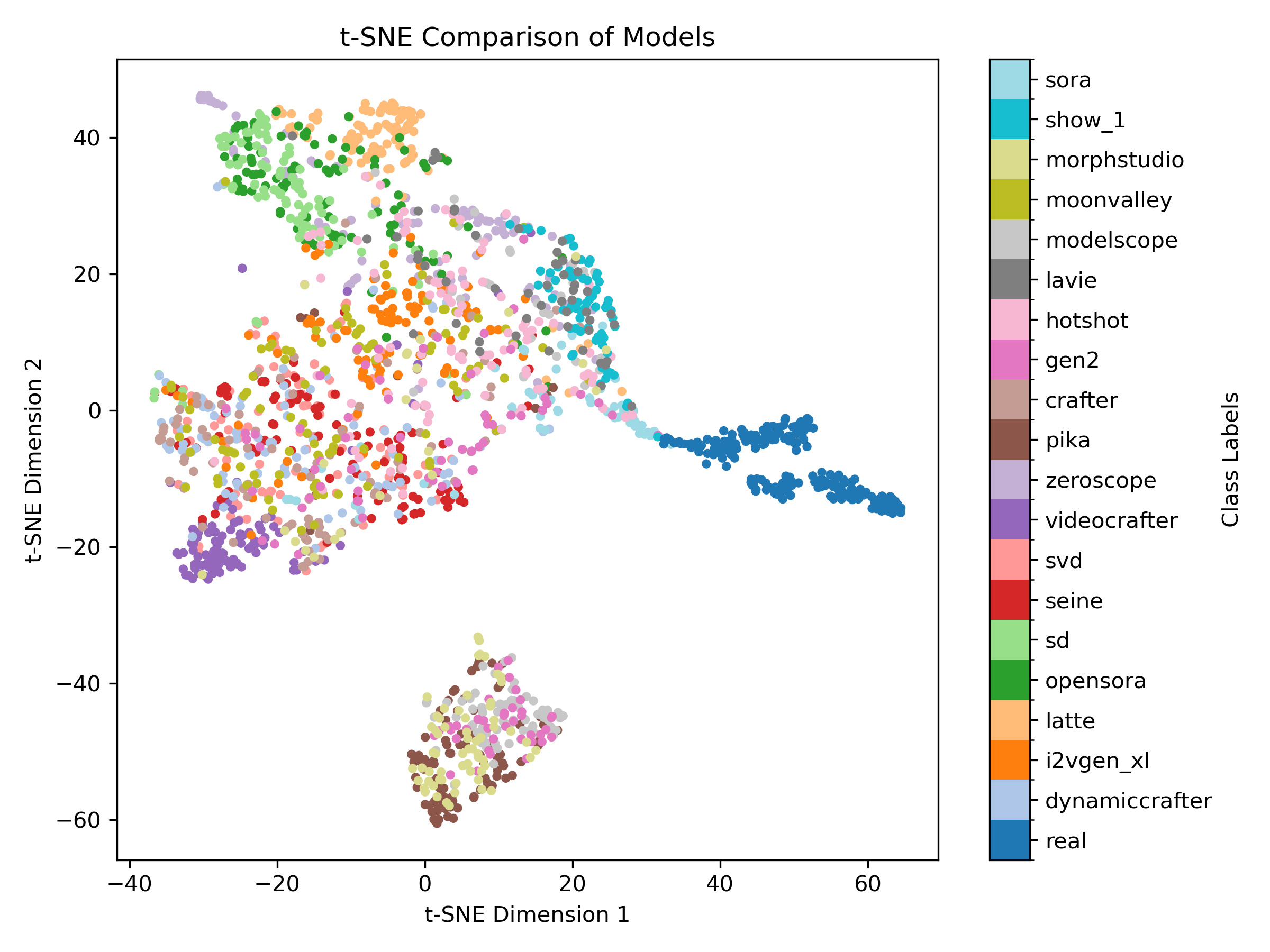}}
    \subfloat[Semi-HNM]{\includegraphics[width=0.33\columnwidth]{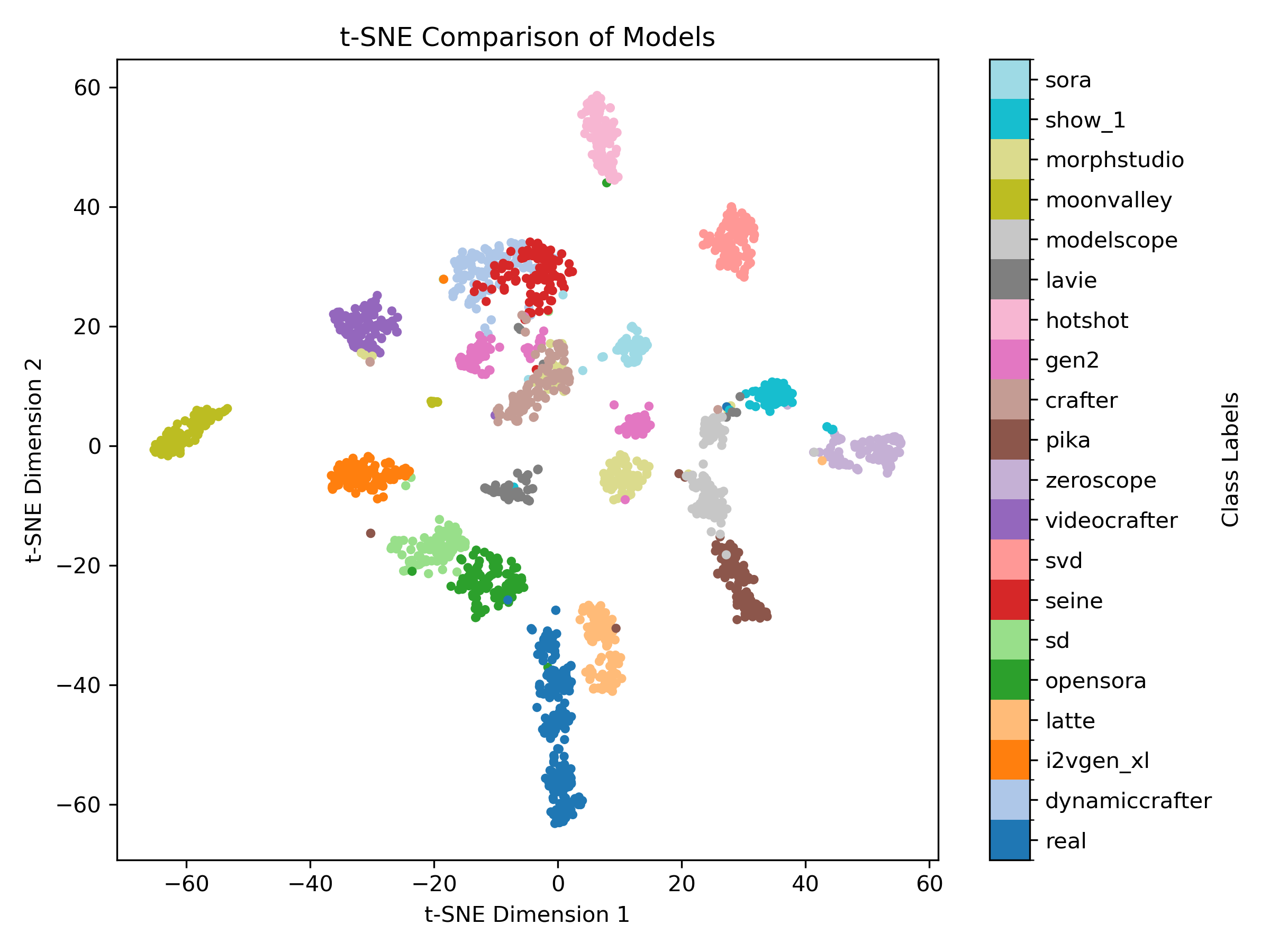}}
    \subfloat[HNM]{\includegraphics[width=0.33\columnwidth]{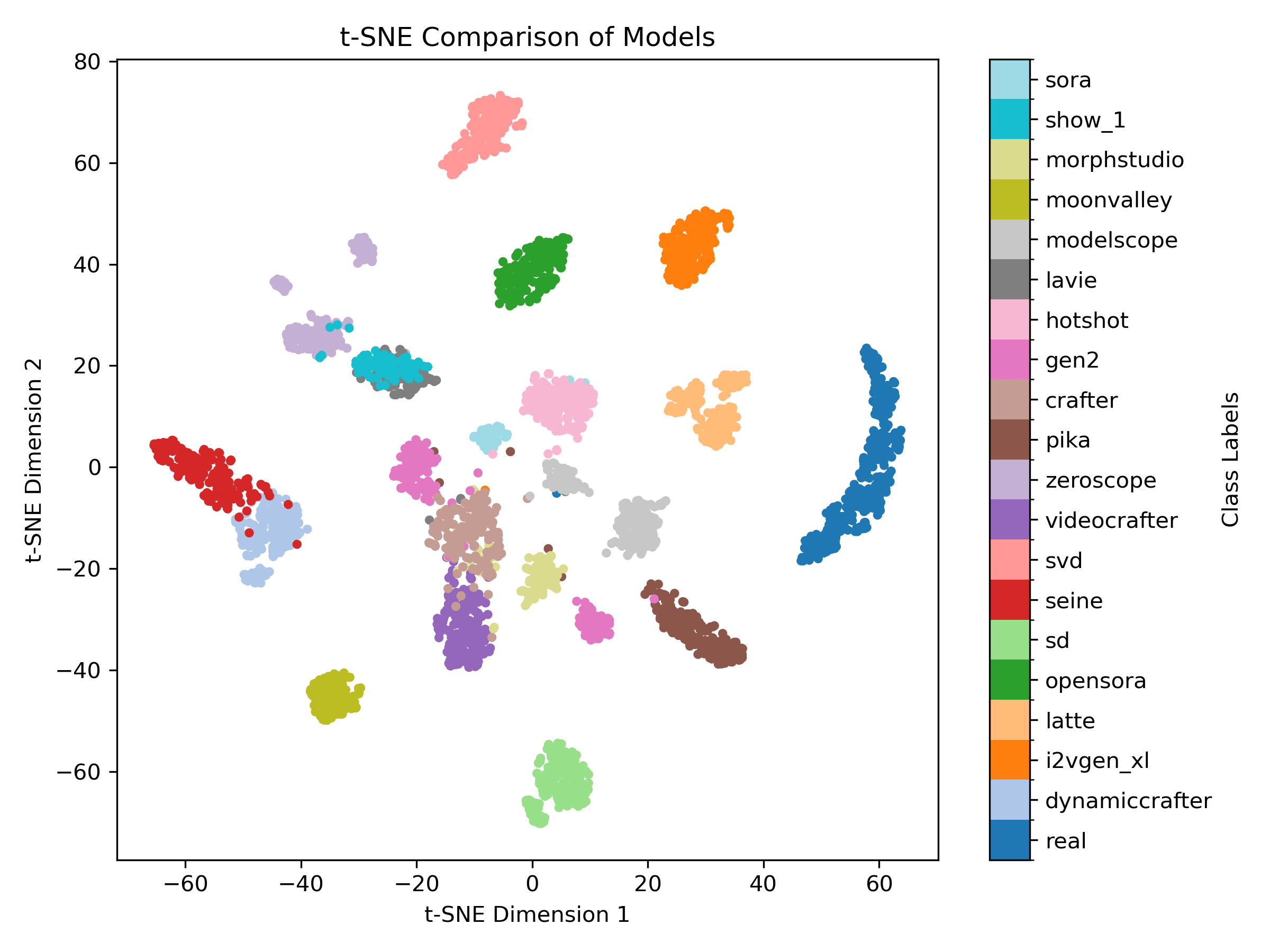}}
    \vspace{-1em}
    \caption{t-SNE visualization of \methodname{} on the \generatorlevel{} attribution task with different loss functions.}
    \vspace{-1em}
    \label{fig:tsne_generator_losses}
\end{figure}

The t-SNE analysis for the \generatorlevel{} attribution task in Fig. \ref{fig:tsne_generator_losses} highlights the impact of different loss functions on \methodname{}’s ability to learn discriminative embeddings. It demonstrates that HNM is highly effective in enforcing discriminative representations for the most fine-grained generator attribution.

\begin{figure}
    \centering
    \includegraphics[width=0.95\columnwidth]{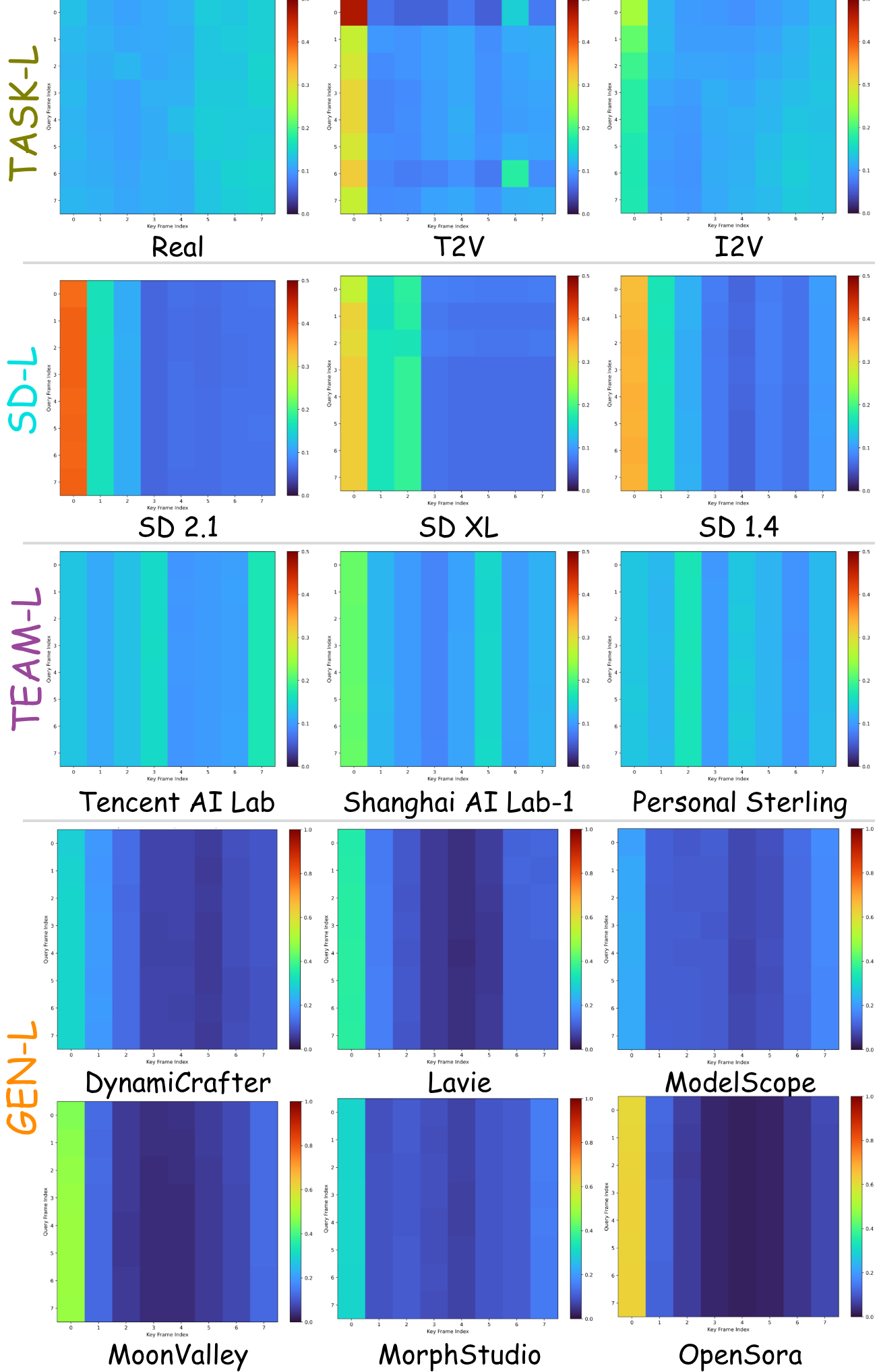}
    \vspace{-1em}
    \caption{\temporalsign{s} for classes in the different attribution levels.}
    \label{fig:temporal_attn}
    \vspace{-0.5em}
\end{figure}

\noindent
\textbf{\underline{\temporalsign{s} Analysis}:} \temporalsign{s} reveal how \methodname{} uses temporal cues to distinguish AI-video sources. These signatures, visualized in Fig. \ref{fig:temporal_attn}, are derived by averaging frame-to-frame attention across a large number of videos per class.

% Intra-Class Consistency \& Inter-Class Discriminability: 
Stable and unique \temporalsign{s} emerge for each class in Fig. \ref{fig:temporal_attn}. Despite content variations, videos from the same source yield consistent signatures, indicating shared temporal artifacts. Crucially, these signatures are visually distinct between classes validating \methodname{}'s ability to capture and differentiate based on class-specific temporal inconsistencies.

% Insights from Unseen Generators: 
As shown in Fig. \ref{fig:teaser}(a), even completely unseen generators produce unique and discernible \temporalsign{s}, distinct from training classes and each other. This suggests \methodname{} learns fundamental temporal characteristics of synthetic generation, beyond just memorizing training patterns. The ability to produce novel signatures for unknown sources indicates strong potential for open-set recognition, allowing \methodname{} to flag content from new generators, which is vital for real-world deployment. In essence, \temporalsign{s} demonstrate that \methodname{} keys in on subtle yet consistent generator-specific temporal fingerprints, enabling accurate and interpretable source attribution.

\noindent
\textbf{\underline{Ablation Results}:} We evaluate the performance of the \methodname{} framework on the \generatorlevel{} attribution task using different loss functions, including CE-loss and contrastive objectives with semi-hard and hard negative mining strategies. Fig. \ref{fig:tsne_generator_losses} (quantitative results in the supplementary) show the superiority of the HNM loss for \generatorlevel{} source attribution tasks. Table~\ref{tab:generator_level} further presents quantitative comparisons under various training regimes: single-stage training with $0.5\%$ labeled samples, single-stage training with $80\%$ of the dataset, and our proposed two-stage training framework utilizing $0.5\%$ labeled data. Across all settings, the contrastive objective with HNM consistently surpasses the CE loss baseline, demonstrating its effectiveness for fine-grained generator-level attribution. We also evaluate the effect of varying the number of samples in second-stage training on source attribution performance, finding that higher sample counts lead to improved results. More details are provided in the supplementary.
\section{Conclusion}\label{sec:conclusion}
We introduced \methodname{}, the first comprehensive framework designed for the critical task of multi-granular source attribution of AI-generated videos, moving beyond inadequate binary detection. By combining a novel video transformer with features from a vision foundation model and a data-efficient two-stage contrastive training strategy, \methodname{} achieves state-of-the-art performance across five attribution levels, from binary to fine-grained generator ID, even with only 0.5\% labeled data and in cross-dataset setups. Our introduction of Temporal Attention Signatures (\temporalsign{s}) provides novel interpretability, visually explaining why generators are distinguishable. \methodname{} establishes a robust benchmark for AI video provenance, offering crucial capabilities for digital forensics and the responsible governance of generative AI.\\

\noindent
\textbf{Acknowledgements:} This work was supported in part by funding from YouTube (Google LLC).

{
    \small
    \bibliographystyle{ieeenat_fullname}
    \bibliography{main}
}
% \newpage
% \appendix
% \input{sections/appendix}

\end{document}